\documentclass{article}

\usepackage{PRIMEarxiv}

\usepackage[utf8]{inputenc} 
\usepackage[T1]{fontenc}    
\usepackage{hyperref}       
\usepackage{url}            
\usepackage{booktabs}       
\usepackage{amsfonts}       
\usepackage{nicefrac}       
\usepackage{microtype}      
\usepackage{lipsum}
\usepackage{fancyhdr}       
\usepackage{graphicx}       
\graphicspath{{media/}}     
\usepackage{adjustbox}
\usepackage{amsmath}
\usepackage{amssymb}
\usepackage{mathtools}
\usepackage{amsthm}
\usepackage{multirow} 
\usepackage{pifont}   
\usepackage[numbers]{natbib}  
\newcommand{\cmark}{\ding{51}} 
\newcommand{\xmark}{\ding{55}} 
\usepackage[table]{xcolor}
\usepackage{wrapfig}
\usepackage{float}
\title{Multi-Level Bidirectional Biomimetic Learning for EEG-Based Visual Decoding}

\author{
Jingtao Liu$^{1}$, \quad
Peiliang Gong$^{1}$, \quad
Chuhang Zheng$^{2}$, \quad
Yiheng Liu$^{1}$, \quad
Qi Zhu$^{1}$
\\[6pt]
$^{1}$ Department of Artificial Intelligence, Nanjing University of Aeronautics and Astronautics, Nanjing, China \\
$^{2}$ Department of Electrical and Information Engineering, Tianjin University, Tianjin, China \\[6pt]
Corresponding author: Qi Zhu \textit{zhuqi@nuaa.edu.cn}
}

\begin{document}
\maketitle

\begin{abstract}
    EEG-based visual neural decoding aims to align neural responses with visual stimuli for tasks such as image retrieval. However, limited paired data and a fundamental mismatch between high-fidelity digital images and biological visual perception—distorted by retinotopic mapping and subject-specific neuroanatomy—severely impede cross-modal alignment. To address this, we propose \textbf{MB$^2$L}, a Multi-Level Bidirectional Biomimetic Learning framework that incorporates structured physiological inductive biases into representation learning. Specifically, we propose {Adaptive Blur with Visual Priors} to mitigate perceptual-structural mismatch by reweighting visual inputs according to retinotopic priors. We further propose {Biomimetic Visual Feature Extraction} to learn multi-level visual representations consistent with hierarchical cortical processing, enhancing subject-invariant encoding. These modules are jointly optimized via {Multi-level Bidirectional Contrastive Learning}, which aligns EEG and visual features in a shared semantic space through bidirectional contrastive objectives. Experiments show MB$^2$L achieves \textbf{80.5\%} Top-1 and \textbf{97.6\%} Top-5 accuracy on zero-shot EEG-to-image retrieval, significantly outperforming prior methods and demonstrating strong generalization across subjects and experimental settings.
\end{abstract}

\section{Introduction}
In the human perceptual system, vision is widely regarded as the dominant sensory modality \cite{hutmacher2019there,jiayu2025neural}, serving as a primary entry point for investigating human perception, cognition, and behavior. This makes understanding the neural processing and response mechanisms underlying visual perception has long been a central focus of neuroscience research. To achieve this, a range of brain imaging techniques have been employed to study visual neural signal decoding, including electroencephalography (EEG) \cite{zheng2021attention, singh2023eeg2image}, magnetoencephalography (MEG) \cite{song2023decoding, benchetrit2023brain}, and functional magnetic resonance imaging (fMRI) \cite{liu2022mind, fang2023alleviating}. Among these modalities, EEG has emerged as a particularly prevalent approach in recent years, owing to its millisecond-level temporal resolution, relative ease of acquisition, and cost-effectiveness, making it especially suitable for capturing the rapid dynamics of human visual processing.

\begin{figure*}[ht]
  \vskip 0.2in
  \begin{center}
    \centerline{\includegraphics[width=\textwidth]{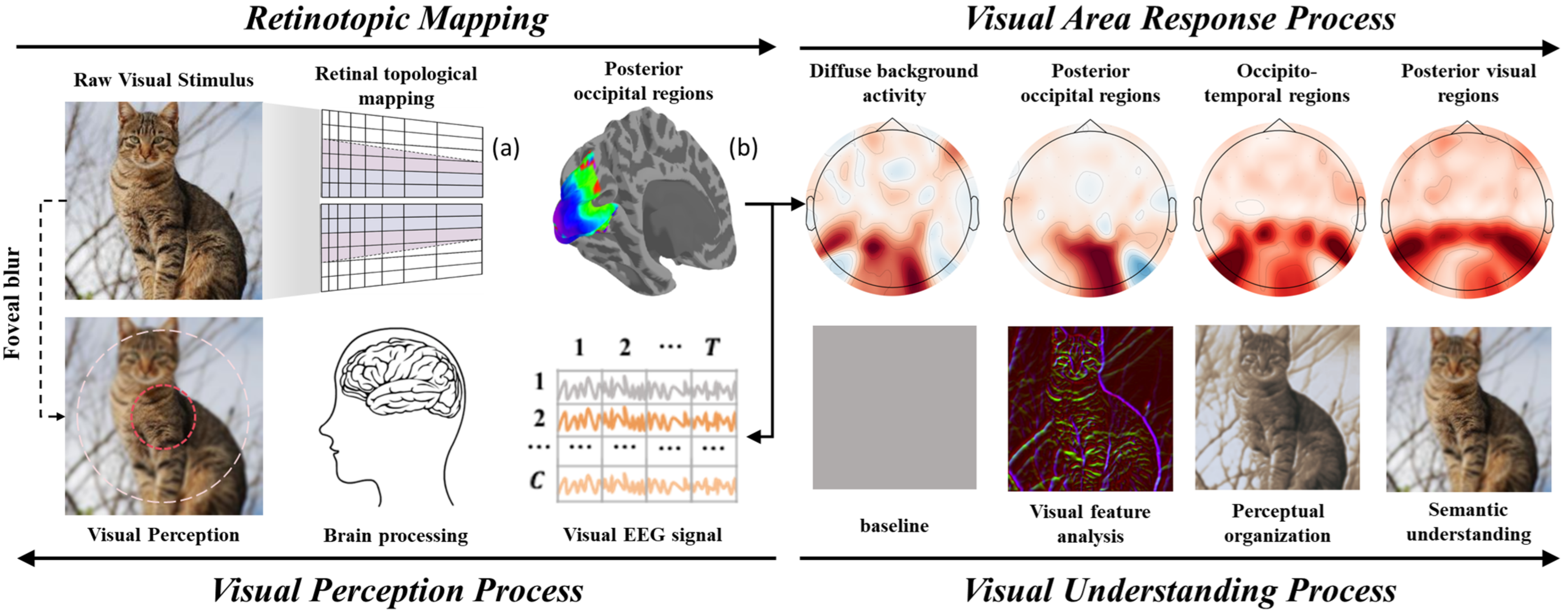}}
    \caption{Schematic of visual processing and neural responses. Left: topographic mapping of visual stimuli in the retina and visual cortex, with panels (a) and (b) reproducing classic studies \cite{ribeiro2025human}. Due to greater cortical resources allocated to foveal input, perception exhibits a fovea-centered blur, differing from the original stimulus. Right: staged neural processing of visual understanding, sequentially involving background-related diffuse activity, low-level feature extraction in the posterior occipital cortex, perceptual organization in the occipitotemporal pathway, and high-level semantic representation in occipital regions, with each stage corresponding to distinct brain activations and visual representations.
    }
    \label{dongji}
  \end{center}
\end{figure*}
EEG visual neural signal decoding aims to establish semantic alignment between the two modalities of visual stimuli and neural signals, enabling retrieval and reconstruction of visual stimuli. \textbf{\textit{However, most existing methods do not adequately account for the inherent modality-specific differences between EEG and visual signals, which manifest in two aspects: information structure differences and individual organizational differences.}} Specifically, structural discrepancies arise from the non-uniform nature of retinotopic mapping. As shown in the left image of Figure~\ref{dongji}, compared to peripheral vision, a larger cortical area is dedicated to central vision \cite{ribeiro2025human}. This results in a foveal blur structure that radiates outward from the fovea, causing information redundancy in the original visual stimuli, particularly in the peripheral regions, relative to neural signals. Furthermore, inter-subject variability in retinal topography and visual cortex organization introduces significant heterogeneity (Ribeiro et al., 2025). Despite a shared functional architecture, these subject-specific anatomical nuances cause identical visual stimuli to elicit divergent neural responses across individuals, thereby severely posing the challenge of cross-modal alignment. Meanwhile, due to the scarcity of paired EEG–image data, relying solely on contrastive learning strategies often struggles to fully exploit the deep semantic associations between the two modalities, thus limiting overall alignment performance. Therefore, it is necessary to introduce physiological prior knowledge during the modeling process and impose physiologically consistent structural constraints, in order to construct a visual neural decoding framework that better aligns with the true mechanisms of human visual processing.

Based on this perspective, we propose MB$^2$L, a \textbf{M}ulti-Level \textbf{B}idirectional \textbf{B}iomimetic \textbf{L}earning framework that models individual-specific neural responses by incorporating physiological priors and contrastive learning for efficient and accurate decoding. Unlike prior methods that focus on empirical alignment, MB$^2$L is the first to jointly model retinotopic structure, cortical hierarchy, and bidirectional cross-modal alignment under a unified biologically constrained framework. Specifically, to address information structure differences between visual stimuli and neural signals, we introduce Adaptive Blur with Visual Priors, which leverages the fovea-to-periphery blur in human perception, incorporates a learnable functional prior, and captures individual variability in retinal topography through adaptive optimization, mitigating peripheral redundancy. Inspired by the hierarchical visual cortex (right panel of Figure~\ref{dongji}), we propose Biomimetic Visual Feature Extraction module to extract low- and high-level visual features in two stages, guiding channel attention weights according to cortical functional divisions to better align feature learning with human visual processing \cite{csikor2025top}. Furthermore, to connect these modules and reduce modality heterogeneity, we adopt Multi-level Bidirectional Contrastive Learning to project features from both modalities into a shared semantic space, enhancing cross-modal interaction.

In summary, MB$^2$L presents a physiologically plausible and efficient visual neural decoding framework. Our main contributions are summarized as follows:
\begin{itemize}
    \item We propose an \textbf{\textit{adaptive blur module with visual priors}}, leveraging shared features of retinal topography to incorporate physiologically grounded priors. It reduces modality discrepancies between visual stimuli and EEG signals while capturing individual-specific retinal structures, enhancing cross-modal alignment.
    \item We introduce a \textbf{\textit{biomimetic visual feature extraction module}}, simulating hierarchical processing of the human visual system under physiologically plausible constraints. By incorporating biologically informed channel-attention priors, it efficiently extracts low- and high-level visual features while reflecting structured cortical processing.
    \item We propose a \textbf{\textit{multi-level bidirectional contrastive learning strategy}}, projecting multi-level EEG and visual features into a shared semantic space to effectively alleviate modality heterogeneity and further strengthen cross-modal correspondence.
    \item Extensive experiments consistently demonstrate the superior performance and generalization of our proposed framework. For zero-shot brain-to-image retrieval, it achieves an \textbf{80.5\%} Top-1 and \textbf{97.6\%} Top-5 accuracy. Furthermore, when adopting Gaussian noise as the image processing strategy, our method achieves a Top-1 accuracy of \textbf{86.2\%}.
\end{itemize}

\section{Related Work}
\subsection{Visual Neural Decoding}
Visual neural decoding \cite{miyawaki2008visual,spampinato2017deep, scotti2024mindeye2, fu2025brainvis} aims to uncover the mapping between visual stimuli and neural responses, enabling interpretable modeling and reconstruction of visual information. With its millisecond-level temporal resolution, EEG has become one of the most commonly used neural signal modalities for studying and decoding dynamic visual processing \cite{zheng2021attention, singh2023eeg2image}.

In recent years, various methods have been proposed for EEG-based visual decoding, most of which are built upon unidirectional contrastive learning frameworks \cite{chen2024visual, du2023decoding, wu2025bridging}. These approaches typically align EEG-derived representations with the embedding space of large-scale vision–language models (e.g., CLIP \cite{pmlr-v139-radford21a}) to achieve cross-modal semantic correspondence.Although recent studies have begun to explore structural aspects of the data, such as the information density gap between EEG and images \cite{wu2025bridging} or hierarchical structures in visual representations \cite{liu2025vieeg}, the internal structural modeling of EEG signals remains relatively underexplored. In particular, the joint modeling of individual perceptual differences (e.g., visual perceptual characteristics) and the hierarchical organization of EEG channels still leaves room for further investigation.
\subsection{Multi-modal Contrastive Learning}
Multimodal contrastive learning has been a long-standing research focus and has achieved substantial progress across domains. Its core goal is to map heterogeneous data, such as images and text, into a shared latent space using large-scale paired semantics and contrastive objectives \cite{schuhmann2022laion}. Models such as CLIP \cite{pmlr-v139-radford21a}, BLIP \cite{li2022blip}, and ALIGN \cite{Jia2021Scaling} leverage massive training data and designed encoders to achieve strong zero-shot retrieval and classification performance, while maintaining generalization to downstream tasks. These successes have further fueled interest in multimodal contrastive learning.

However, specific modality combinations such as video–text \cite{bain2021frozen}, 3D–language \cite{xue2023ulip}, and EEG–image still face challenges in real-world scenarios. These tasks often lack large-scale, high-quality paired data and exhibit pronounced cross-modal heterogeneity. For EEG–image alignment, the complexity and noise of neural signals further hinder effective alignment, limiting model generalization, particularly in zero-shot settings. Therefore, alleviating data scarcity and improving accuracy and consistency of cross-modal alignment remain key open problems in multimodal representation learning.

\subsection{Visual Prior and Robust Perception Modeling}
Visual perception is not a direct mapping of the original visual stimuli but rather a robust information integration process that relies on intrinsic priors in complex and uncertain environments \cite{moreno2011bayesian, bowers2023deep}. In natural visual scenes, degradation factors such as blur, noise, and resolution variations are ubiquitous. However, the human visual system is still able to maintain stable perception, indicating the crucial role of multi-scale feature integration and perceptual priors in visual processing.

Existing studies \cite{ribeiro2025human,jang2024improved} have shown that different forms of visual degradation selectively modulate feature representations, with visual cortical areas exhibiting distinct sensitivity to spatial frequency and clarity: low-level regions emphasize local structures, whereas high-level regions exhibit semantic invariance. This suggests that visual degradation reflects hierarchical perceptual influences rather than mere information loss. However, most EEG-based visual decoding and cross-modal alignment approaches \cite{wu2025bridging, zhang2025neurobridge} treat visual degradation primarily as a data augmentation strategy, with limited investigation into how perceptual priors shape neural representations. Effectively integrating visual priors into EEG–image alignment to address perceptual inconsistencies remains an open challenge.
\section{Method}
\begin{figure*}[ht]
  \vskip 0.2in
  \begin{center}
    \centerline{\includegraphics[width=\textwidth]{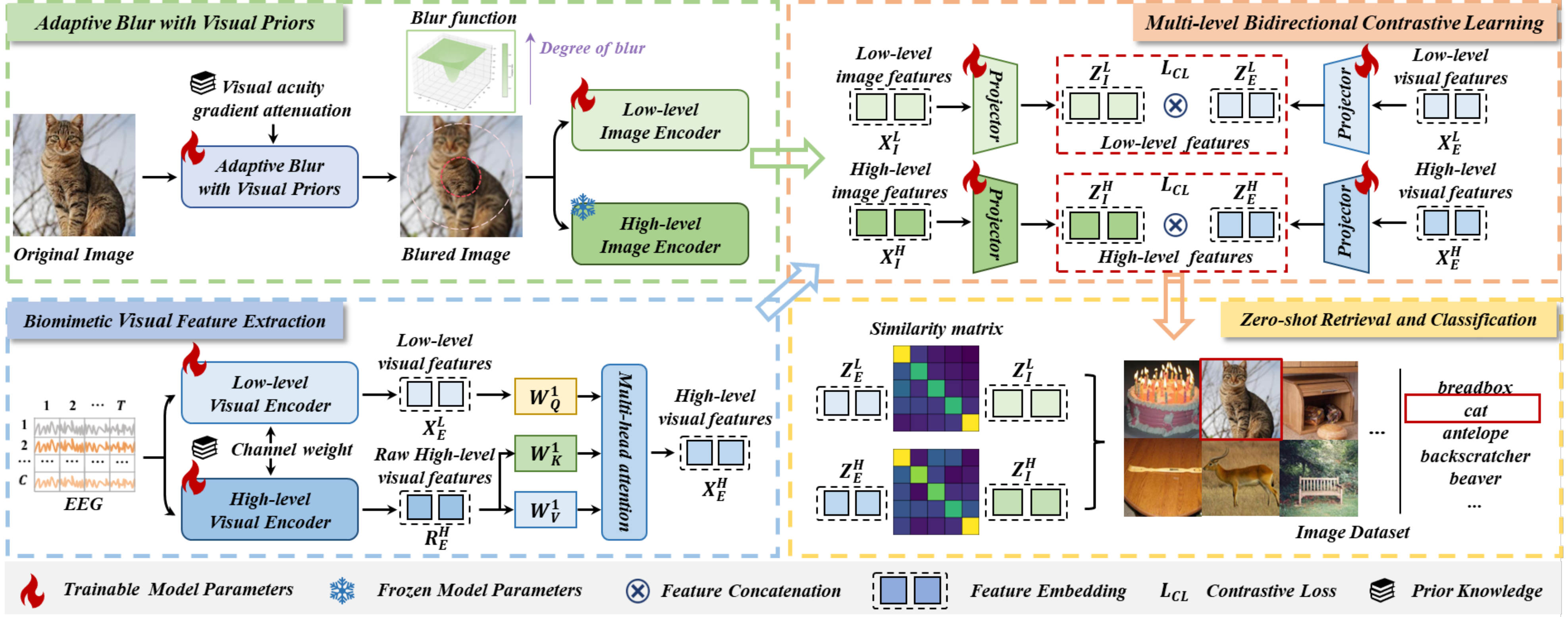}}
    \caption{Overall framework of MB$^2$L.(1) Adaptive Blur with Visual Priors (top-left): The original image undergoes biomimetic blurring, then hierarchical visual features are extracted via low- and high-level encoders; (2) Biomimetic Visual Feature Extraction (bottom-left): EEG signals are split by a channel-weighted layer, then encoded into hierarchical features through cross-attention; (3) Multi-level Bidirectional Contrastive Learning (right): Cross-modal features are projected and aligned via contrastive loss, enabling zero-shot retrieval and classification.
    }
    \label{method}
  \end{center}
\end{figure*}
We propose MB$^2$L, a multi-Level bidirectional biomimetic learning framework composed of three main modules: Adaptive Blur with Visual Priors (ABVP), Biomimetic Visual Feature Extraction (BVFE), and Multi-level Bidirectional Contrastive Learning (MBCL). The overall architecture of the proposed framework is illustrated in the figure \ref{method}.

During training, paired EEG–image data are used to learn modality-specific representations, while contrastive learning explicitly aligns matched pairs across modalities and removes mismatched pairs in the shared embedding space. During inference, the trained encoder and mapper efficiently perform zero-shot classification and image retrieval based on EEG signals alone, demonstrating the effectiveness of the proposed framework for EEG-based visual decoding.

\subsection{Adaptive Blur with Visual Priors}
Due to the constraints of retinal topography, visual perception exhibits increasing blur from the fovea to the periphery, leading to peripheral information loss and suggesting that raw visual stimuli contain redundant data relative to brain signals. To address this structural mismatch, we propose an adaptive blur method that incorporates visual priors and learns the blurring effect in a data-driven manner, thereby enhancing the efficiency of cross-modal alignment.

Based on the characteristics of the dataset acquisition paradigm, in which participants were instructed to fixate on a red dot at the center of each image, we treat the image center as the visual center. Guided by visual priors, we construct a blur function whose parameters are learned to achieve adaptive and spatially varying blurring. Following prior studies \cite{wu2025bridging}, Gaussian blur is adopted as the image information attenuation method. Specifically, we first apply Gaussian blurring to the original image to obtain a uniformly blurred image:

$$I_{blur}(i,j) = \sum^{k}_{m=-k}\sum^{k}_{n=-k}I(i-m,j-n)G_{\sigma}(m,n)$$

where $k = \lfloor r_{kernel}/2 \rfloor$, $r_{kernel}$ represents the radius of the Gaussian kernel,$ \ I$ represents the original image, and $I_{blur} $ represents the generated blurred image.

$$G_{\sigma}(m,n) = \frac{1}{Z}exp(-\frac{m^2+n^2}{2\sigma^2})$$

where $Z$ denotes the normalization constant and $\sigma$ the standard deviation, which controls the overall degree of blurriness. The procedure for generating blurred images is as follows:

$$\hat{I}_{visual} = w I_{blur}+(1-w)I$$

where $w(i,j) \in [0,1]$ represents the fusion weight. Physiologically, the visual resolution is the highest in the fovea region of the retina; as the eccentricity increases, beyond the radius of the fovea, the visual resolution rapidly decreases, and the corresponding perceptual blurriness increases rapidly; in the peripheral areas far from the center, this blurriness increase tends to be stable:

$$w(r)=\sigma_{act}(k(r-r_0))$$

where $\sigma_{act}$  represents the activation function, $k$ is the blur coefficient, $r_0$ represents the radius of the fovea, and $r(i,j)$ represents the Euclidean distance between the pixel $(i,j)$ and the visual center, which is the origin of the fovea. In our setup, k and $r_0$ are learnable parameters, and the optimal blur scheme is found through contrast learning between the two modalities.
\subsection{Biomimetic Visual Feature Extraction}
Human visual processing is hierarchical: the retina converts light into neural signals, relayed via the lateral geniculate nucleus to early visual areas, where low-level features are extracted and progressively processed by higher cortical regions such as the inferior temporal lobe, generating high-level features for object and scene recognition. Inspired by this, we propose a progressive visual feature extraction framework incorporating channel priors.

We plan to decouple low-level and high-level visual features from EEG signals. First, we introduce a simple channel attention layer to allocate attention weights to the signals from each channel. Priors are incorporated based on electrode positions and the functions of the visual cortex, setting the initial weights, which can be formulated as:
$$E^H=w^HE^R,E^L=w^LE^R$$
where $E^H$ represents the high-level visual bias signal, $E^L$ represents the low-level visual bias signal, $E^R$ represents the original signal, $w^H$ and $w^L$ represent the corresponding learnable weights, and they are initialized according to physiological reality.

Then, we followed the processing flow of visual neural signals and extracted the low-level visual features $Z_E^L$ from $E^L$. Using this as a guide, we employed the cross-attention mechanism to extract the high-level visual features $Z_E^H $from $E^H$. The specific process can be formulated as:
$$Z_E^L = f^L(E^L), Z_E^H = CrossAttn(Z_E^L, f^H(E^H))$$
$$CrossAttn(X,Y)=softmax(\frac{(XW_Q)(YW_K)^T}{\sqrt{d}})(YW_V)$$
Among them, $W_Q$, $W_K$, and $W_V$ are learnable weight matrices used to map the input into "query vector", "key vector" and "value vector", and $\sqrt d$ is the scaling factor.

This biomimetic visual feature extraction framework is inspired by the progressive processing mechanism of the human visual system and incorporates physiological channel priors. It not only follows the hierarchical principles of biological visual cognition but also enhances the specificity and semantic consistency of EEG visual features, providing a more cognitively coherent and logical feature representation basis for subsequent EEG-based visual information recognition and understanding.
\subsection{Multi-level Bidirectional Contrastive Learning}
In EEG-based visual decoding, neural signals capture the hierarchical stages of visual processing, from early sensory encoding to higher-level semantic abstraction. Directly mapping EEG representations to a single image embedding space not only fails to reflect this hierarchical structure but also implicitly assumes alignment in scale and semantic level between EEG and image representations, which is unsupported by neurophysiological and representation learning evidence.

To address this, we propose MBCL, which uses a trainable shallow Resnet and a frozen pre-trained visual model to decouple low-level and high-level image features from the original visual stimulus, corresponding to the low-level and high-level visual features of the EEG representation, and maps them respectively to the corresponding trainable semantic spaces. It achieves effective cross-modal alignment by learning a modality-independent latent space. The projection network can be flexibly adjusted to adapt to different tasks and datasets.The projection process is formulated as:
$$Z_E^L = p_E^L(X_E^L),Z_E^H = p_E^H(X_E^H),Z_I^L = p_I^L(X_I^L),Z_I^H = p_I^H(X_I^H) $$
where $p$ represents the projection network, $Z$ represents the mapping embedding of EEG representation and image representation in the latent space, the subscript indicates the modality to which it belongs, $I$ represents the image modality, $E$ represents the EEG modality, the superscript $L$ indicates low-level features, and $H$ indicates high-level features.

We use bidirectional contrastive learning to synchronously train the EEG encoder, the shallow Resnet, and four projection networks. We optimize the entire network by bringing the matching pairs closer and pushing the non-matching pairs apart. The loss settings are as follows:
\[
\begin{aligned}
L_{total} &= \alpha^L L(Z_I^L,Z_E^L)+ \alpha^H L(Z_I^H,Z_E^H)\\
\mathcal{L}(Z_I,Z_E) &= \frac{1}{2N} \sum_{j=1}^{N} \left( \mathcal{L}_{\text{I→E}}(j) + \mathcal{L}_{\text{E→I}}(j) \right) \\
\mathcal{L}_{\text{I→E}}(j) &= -\log \frac{\exp\left( \mathrm{sim}(z_{I,j}, z_{E,j}) / \tau \right)}{\sum_{z^-_{E,j} \in \bar{\mathcal{Z}}_{E,j}} \exp\left( \mathrm{sim}(z_{I,j}, z^-_{E,j}) / \tau \right)} \\
\mathcal{L}_{\text{E→I}}(j) &= -\log \frac{\exp\left( \mathrm{sim}(z_{E,j}, z_{I,j}) / \tau \right)}{\sum_{z^-_{I,j} \in \bar{\mathcal{Z}}_{I,j}} \exp\left( \mathrm{sim}(z_{E,j}, z^-_{I,j}) / \tau \right)}
\end{aligned}
\]

where $\alpha^L$ and $\alpha^H$ represent the loss weights, $Z_{I,j} \in Z_I$ and $Z_{E,j} \in Z_E$ represent image and EEG representations, while $z_{I,j}^-$ and $Z_{E,j}^-$ are their negative sets. $N$ is the number of data pairs, $\tau$ is the temperature coefficient, and $sim(\cdot,\cdot)$ uses cosine similarity.

Unlike traditional learning based on single and fixed image representations, we consider and simulate the hierarchical structure of visual processing and construct a learnable shared latent space, which effectively alleviates heterogeneous representation issues between the two modalities and further improves the robustness and accuracy of visual decoding performance.
\section{Experiments and Results}
\subsection{Datasets and Implementation Details}
\textbf{THINGS-EEG} \cite{gifford2022large} includes data from 10 participants collected using Rapid Serial Visual Presentation (RSVP) paradigm \cite{grootswagers2019representational,intraub1981rapid,keysers2001speed}. The training set consists of 1,654 concepts (10 images each), with each participant completing 4 equivalent trials. The test set contains 200 concepts (1 image each), with each participant completing 80 equivalent trials. Data preprocessing follows methods established in previous studies \cite{song2023decoding,wu2025bridging}, where repeated trials are averaged to improve signal-to-noise ratio and data reliability. Ultimately, each participant has 16,540 training samples and 200 test samples.

\textbf{THINGS-MEG} \cite{hebart2023things} was collected from four participants, consisting of recordings from 271 MEG channels. The training set comprises 1,654 concepts, each with 12 images presented once, while the test set contains 200 concepts, each with a single image repeated 12 times. Following the settings in existing studies \cite{song2023decoding, wu2025bridging}, repeated trials under the same stimulus conditions were averaged to enhance the signal-to-noise ratio and improve data reliability. 

\textbf{EEG-Encoder}. We adopt EEGProject from UBP \cite{wu2025bridging} as the default EEG encoder in our framework. We further include representative EEG encoders for comparison, including Shallownet \cite{schirrmeister2017deep}, EEGNet \cite{lawhern2018eegnet}, and TSConv \cite{song2023decoding}, to evaluate generalization.

\textbf{Image-Encoder}. We use ResNet-18/34/50/101/152 \cite{he2016deep} to extract hierarchical visual representations and assess robustness across network depths.

Additional dataset details are provided in Appendix~\ref{datasets}.
\subsection{Comparison with Baselines}
\vspace{-15pt}

\begin{table*}[htbp]
  \caption{Top-1 and Top-5 accuracy (\%) for 200-way zero-shot retrieval on THINGS-EEG}
  \label{tab:acc1}
  
  \centering
  \adjustbox{max width=\linewidth}{
  \begin{tabular}{lccccccccccccccccccccccc}
    \toprule
    \multirow{2}{*}{Method} & \multicolumn{2}{c}{Subj. 1} & \multicolumn{2}{c}{Subj. 2} & \multicolumn{2}{c}{Subj. 3} & \multicolumn{2}{c}{Subj. 4} & \multicolumn{2}{c}{Subj. 5} & \multicolumn{2}{c}{Subj. 6} & \multicolumn{2}{c}{Subj. 7} & \multicolumn{2}{c}{Subj. 8} & \multicolumn{2}{c}{Subj. 9} & \multicolumn{2}{c}{Subj. 10} & \multicolumn{2}{c}{Avg} \\
    \cmidrule(lr){2-3} \cmidrule(lr){4-5} \cmidrule(lr){6-7} \cmidrule(lr){8-9} \cmidrule(lr){10-11} \cmidrule(lr){12-13} \cmidrule(lr){14-15} \cmidrule(lr){16-17} \cmidrule(lr){18-19} \cmidrule(lr){20-21} \cmidrule(lr){22-23}
    & top-1 & top-5 & top-1 & top-5 & top-1 & top-5 & top-1 & top-5 & top-1 & top-5 & top-1 & top-5 & top-1 & top-5 & top-1 & top-5 & top-1 & top-5 & top-1 & top-5 & top-1 & top-5 \\
    \midrule
    \multicolumn{23}{c}{\textbf{Intra-subject : train and test on one subject}} \\
    \midrule
    BraVL & 6.1 & 17.9 & 4.9 & 14.9 & 5.6 & 17.4 & 5.0 & 15.1 & 4.0 & 13.4 & 6.0 & 18.2 & 6.5 & 20.4 & 8.8 & 23.7 & 4.3 & 14.0 & 7.0 & 19.7 & 5.8 & 17.5 \\
    NICE  & 13.2 & 39.5 & 13.5 & 40.3 & 14.5 & 42.7 & 20.6 & 52.7 & 10.1 & 31.5 & 16.5 & 44.0 & 17.0 & 42.1 & 22.9 & 56.1 & 15.4 & 41.6 & 17.4 & 45.8 & 16.1 & 43.6 \\
    NICE-SA & 13.3 & 40.2 & 12.1 & 36.1 & 15.3 & 39.6 & 15.9 & 49.0 & 9.8 & 34.4 & 14.2 & 42.4 & 17.9 & 43.6 & 18.2 & 50.2 & 14.4 & 38.7 & 16.0 & 42.8 & 14.7 & 41.7 \\
    NICE-GA & 15.2 & 40.1 & 13.9 & 40.1 & 14.7 & 42.7 & 17.6 & 48.9 & 9.0 & 29.7 & 16.4 & 44.4 & 14.9 & 43.1 & 20.3 & 52.1 & 14.1 & 39.7 & 19.6 & 46.7 & 15.6 & 42.8 \\
    ATM-S & 25.6 & 60.4 & 22.0 & 54.5 & 25.0 & 62.4 & 31.4 & 60.9 & 12.9 & 43.0 & 21.3 & 51.1 & 30.5 & 61.5 & 38.8 & 72.0 & 34.4 & 51.5 & 29.1 & 63.5 & 28.5 & 60.4 \\
    Neural-MCRL  & 27.5 & 64.0 & 28.5 & 61.5 & 37.0 & 69.0 & 35.0 & 66.0 & 22.5 & 51.5 & 31.5 & 61.0 & 31.5 & 62.5 & 42.0 & 74.5 & 30.5 & 59.5 & 37.5 & 71.0 & 32.4 & 64.1 \\
    VE-SDN  & 32.6 & 63.7 & 34.4 & 69.9 & 38.7 & 73.5 & 39.8 & 72.0 & 29.4 & 58.6 & 34.5 & 68.8 & 34.5 & 68.3 & 49.3 & 79.8 & 39.0 & 69.6 & 39.8 & 75.3 & 37.2 & 69.9 \\
    CognitionCapturer & 27.2 & 59.5 & 28.7 & 57.0 & 37.2 & 66.1 & 37.7 & 63.2 & 21.8 & 47.8 & 31.6 & 58.1 & 32.8 & 59.6 & 47.6 & 73.5 & 33.4 & 57.7 & 35.1 & 63.6 & 33.3 & 60.6 \\
    UBP  & \underline{41.2} & \underline{70.5} & \underline{51.2} & \underline{80.9} & \underline{51.2} & \underline{82.0} & \underline{51.1} & \underline{76.9} & \underline{42.2} & \underline{72.8} & \underline{57.5} & \underline{83.5} & \underline{49.0} & \underline{79.9} & \underline{58.6} & \underline{85.8} & \underline{45.1} & \underline{76.2} & \underline{61.5} & \underline{88.2} & \underline{50.9} & \underline{79.7} \\
\rowcolor{gray!20}
\textbf{MB$^2$L(Ours)} & \textbf{82.5} & \textbf{99.0} & \textbf{85.0} & \textbf{99.0} & \textbf{80.0} & \textbf{98.0} & \textbf{67.0} & \textbf{95.0} & \textbf{74.5} & \textbf{95.5} & \textbf{86.5} & \textbf{98.5} & \textbf{78.5} & \textbf{96.0} & \textbf{86.5} & \textbf{99.5} & \textbf{77.5} & \textbf{95.5} & \textbf{86.5} & \textbf{99.5} & \textbf{80.5} & \textbf{97.6} \\
    \midrule
    \multicolumn{23}{c}{\textbf{Inter-subject : leave one subject out for test}} \\
    \midrule
    BraVL & 2.3 & 8.0 & 1.5 & 6.3 & 1.4 & 5.9 & 1.7 & 6.7 & 1.5 & 5.6 & 1.8 & 7.2 & 2.1 & 8.1 & 2.2 & 7.6 & 1.6 & 6.4 & 2.3 & 8.5 & 1.8 & 7.0 \\
    NICE  & 7.6 & 22.8 & 5.9 & 20.5 & 6.0 & 22.3 & 6.3 & 20.7 & 4.4 & 18.3 & 5.6 & 22.2 & 5.6 & 19.7 & 6.3 & 22.0 & 5.7 & 17.6 & 8.4 & 28.3 & 6.2 & 21.4 \\
    NICE-SA  & 7.0 & 22.6 & 6.6 & 23.2 & 7.5 & 23.7 & 5.4 & 21.4 & 6.4 & 22.2 & 7.5 & 22.5 & 3.8 & 19.1 & 8.5 & 24.4 & 7.4 & 22.3 & 9.8 & 29.6 & 7.0 & 23.1 \\
    NICE-GA & 5.9 & 21.4 & 6.4 & 22.7 & 5.5 & 20.1 & 6.1 & 21.0 & 4.7 & 19.5 & 6.2 & 22.5 & 5.9 & 19.1 & 7.3 & 25.3 & 4.8 & 18.3 & 6.2 & 26.3 & 5.9 & 21.6 \\
    ATM-S  & 10.5 & 26.8 & 7.1 & 24.8 & 11.9 & \underline{33.8} & \underline{14.7} & \underline{39.4} & 7.0 & 23.9 & 11.1 & \underline{35.8} & \underline{16.1 }&  \textbf{43.5} & 15.0 & \textbf{40.3} & 4.9 & 22.7 & \underline{20.5} & \underline{46.5} & 11.8 & 33.7 \\
    Neural-MCRL & 13.0 & \underline{31.5} & 12.0 & 30.5 & \underline{14.5} & \textbf{35.5} & 12.5 & 35.5 & \underline{11.5} & 29.0 & \underline{13.5} & 35.5 & 14.0 & \underline{36.0} & \textbf{18.5} & \underline{38.5} & 13.5 & 32.5 & 17.0 & 39.0 & \underline{14.0} & \underline{34.3} \\
    UBP  
& \underline{11.5} & 29.7 & \underline{15.5} & \underline{40.0} & 9.8 &27.0 & 13.0 & 32.3 & 8.8 & \textbf{33.8} & 11.7 & 31.0 & 10.2 & 23.8 & 12.2 & 32.2 & \underline{15.5} & \underline{40.5} & 16.0 & 43.5 & 12.4 & 33.4 \\
\rowcolor{gray!20}
\textbf{MB$^2$L(Ours)} & \textbf{22.5} & \textbf{62.5} & \textbf{30.0} & \textbf{57.0} & \textbf{16.5} & \textbf{35.5} & \textbf{20.5} & \textbf{43.5} & \textbf{12.0} & \underline{32.0} & \textbf{16.5} & \textbf{41.5} & \textbf{21.0} & \textbf{43.5} & \underline{15.5} & 35.0 & \textbf{17.0} & \textbf{42.0} & \textbf{26.5} & \textbf{54.5} & \textbf{19.8} & \textbf{44.7} \\ 
    \bottomrule
  \end{tabular}
  }
\end{table*}

\textbf{Baseline}. We compare our method with several recent and representative EEG-based visual decoding approaches, including BraVL \cite{du2023decoding}, NICE \cite{song2023decoding}, ATM \cite{li2024visual}, Neural-MCRL \cite{li2024neural}, CognitionCapturer \cite{zhang2025cognitioncapturer}, VE-SDN \cite{chen2024visual}, and UBP \cite{wu2025bridging}. Detailed descriptions and implementation details of these baseline methods are provided in Appendix~\ref{sec:comparison_method}.

\begin{wraptable}{R}{0.48\textwidth}
  \centering
  \vspace{-18pt}
  \renewcommand{\arraystretch}{1.0}
  \setlength{\tabcolsep}{3.2pt} 
  \caption{Top-1 and Top-5 accuracy (\%) for 200-way zero-shot retrieval on THINGS-MEG}
  \resizebox{0.48\textwidth}{!}{\begin{tabular}{lcccccccccc}
    \toprule
    \multirow{2}{*}{Method} & \multicolumn{2}{c}{S1} & \multicolumn{2}{c}{S2} & \multicolumn{2}{c}{S3} & \multicolumn{2}{c}{S4} & \multicolumn{2}{c}{Avg} \\
    \cmidrule(lr){2-3} \cmidrule(lr){4-5} \cmidrule(lr){6-7} \cmidrule(lr){8-9} \cmidrule(lr){10-11}
    & top-1 & top-5 & top-1 & top-5 & top-1 & top-5 & top-1 & top-5 & top-1 & top-5 \\
    \midrule
    \multicolumn{11}{c}{\textbf{Intra-subject}} \\
    \midrule
    NICE & 9.6 & 27.8 & 18.5 & 47.8 & 14.2 & 41.6 & 9.0 & 26.6 & 12.8 & 36.0 \\
    NICE-SA & 9.8 & 27.8 & 18.6 & 46.4 & 10.5 & 38.4 & 11.7 & 27.2 & 12.7 & 35.0 \\
    NICE-GA & 8.7 & 30.5 & 21.8 & 56.6 & 16.5 & 49.7 & 10.3 & 32.3 & 14.3 & 42.3 \\
    UBP & \underline{15.0} & \underline{38.0} & \underline{46.0} & \underline{80.5} & \underline{27.3} & \underline{59.0} & \underline{18.5} & \underline{43.5} & \underline{26.7} & \underline{55.2} \\
    \rowcolor{gray!20}
    \textbf{MB$^2$L}& \textbf{19.0} & \textbf{42.0} & \textbf{66.5} & \textbf{88.0} & \textbf{38.0} & \textbf{72.5} & \textbf{21.0} & \textbf{50.5} & \textbf{36.1} & \textbf{69.5} \\
    \midrule
    \multicolumn{11}{c}{\textbf{Inter-subject}} \\
    \midrule
    UBP & \underline{2.0} & \underline{5.7} & \underline{1.5} & \underline{17.2} & \underline{2.7} & \underline{10.5} & \underline{2.5} & \underline{8.0} & \underline{2.2} & \underline{10.4} \\
    \rowcolor{gray!20}
    \textbf{MB$^2$L} & \textbf{4.5} & \textbf{12.5} & \textbf{6.0} & \textbf{17.0} & \textbf{7.5} & \textbf{18.0} & \textbf{3.0} & \textbf{12.5} & \textbf{5.3} & \textbf{15.0} \\
    \bottomrule
  \end{tabular}}
  \label{tab:acc2}
  \vspace{-8pt}
\end{wraptable}
\vspace{0pt}
\textbf{Comparison}. Tables \ref{tab:acc1} and \ref{tab:acc2} present quantitative comparison results of our method with baseline methods on EEG and MEG test sets under the same experimental settings (see Appendix~\ref{Imp details}). In both within-subject and across-subject evaluation settings, our method consistently achieves state-of-the-art performance, demonstrating robust generalization across different subjects. In particular, in the zero-shot brain-to-image retrieval task, our method attains Top-1 and Top-5 accuracies of 80.5\% and 97.6\% on the THINGS-EEG dataset, and 36.1\% and 69.5\% on the THINGS-MEG dataset, respectively. These results highlight the effectiveness of our framework in capturing subject-invariant neural patterns and suggest its potential for scalable neural decoding across diverse participants and recording modalities.

\subsection{Ablation Study on Core Framework Components}

\begin{wraptable}{R}{0.48\textwidth}
  \centering
  \vspace{-18pt}
  \renewcommand{\arraystretch}{1.0}
  \setlength{\tabcolsep}{6.5pt}
  \caption{Ablation study on the core components of MB$^2$L(\%)}
  \begin{tabular}{cccccc}
    \toprule
    \textbf{ABVP} & \textbf{BVFE} & \textbf{MBCL} & \textbf{Top-1} & \textbf{Top-5} \\
    \midrule
    \xmark & \xmark & \xmark & 27.2 & 59.4 \\
    \xmark & \cmark & \xmark & 25.3 & 54.5 \\
    \xmark & \xmark & \cmark & 65.4 & 92.1 \\
    \cmark & \xmark & \cmark & \underline{75.9} & \underline{96.6} \\
    \xmark & \cmark & \cmark & 69.7 & 94.0 \\
    \rowcolor{gray!20}
    \cmark & \cmark & \cmark & \textbf{80.5} & \textbf{97.6} \\
    \bottomrule
  \end{tabular}
  \label{tab:ablation}
  \vspace{-8pt}
\end{wraptable}
\vspace{0pt}

To evaluate each component, ablation experiments were conducted by removing ABVP (eliminating adaptive blurring and using original visual stimuli), BVFE (removing the channel prior and cross-attention module), and MBCL (excluding trainable modules on the image side); partial experiments were omitted due to ABVP’s reliance on MBCL. As shown in Table~\ref{tab:ablation}, all components positively affect performance. While BVFE alone causes a slight decline, its combination with other modules consistently improves results, as MBCL mitigates modality heterogeneity and enables BVFE to contribute to feature learning. Detailed ablation results and further analyses are provided in Appendix~\ref{abl}.

\subsection{Comparative Analysis of Different Image Processing Methods}
\begin{figure}[H]
  \centering
  \begin{minipage}{0.48\textwidth}
    \centering
    \includegraphics[width=\linewidth]{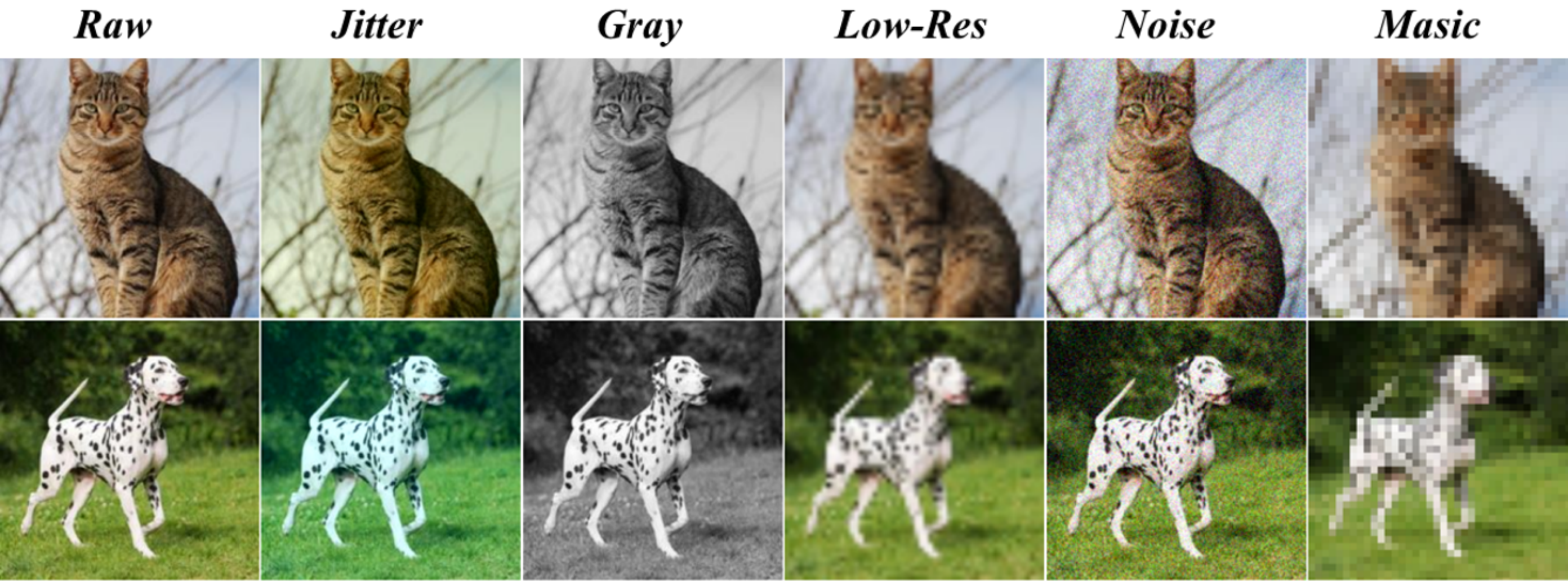}
    \caption{Visualization of representative images processed with different image processing methods used in the experiments.}
    \label{fig:img process}
  \end{minipage}
  \hfill
  \begin{minipage}{0.48\textwidth}
    \centering
    \includegraphics[width=\linewidth]{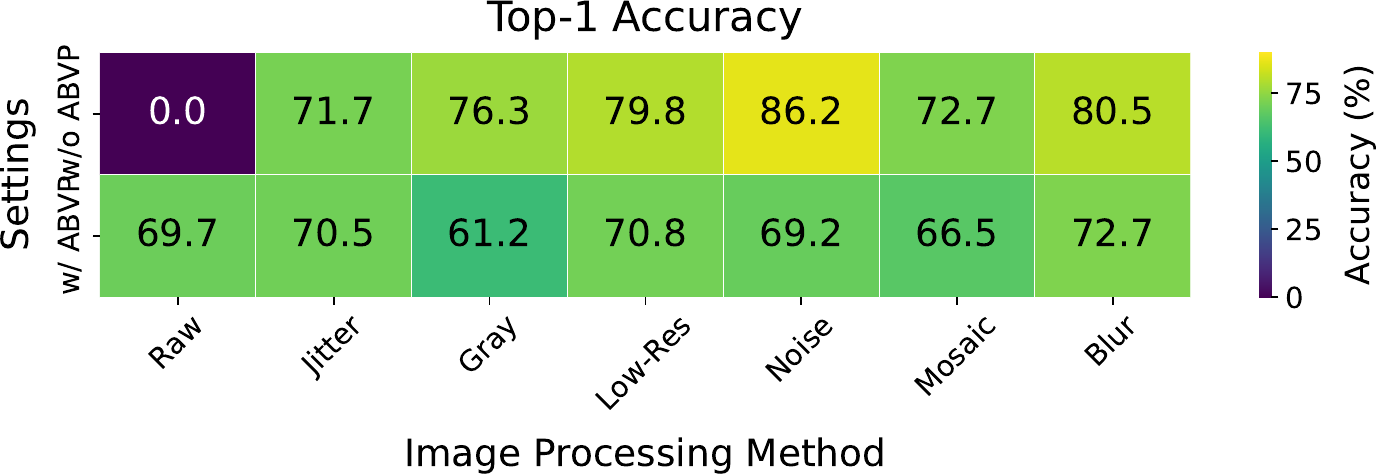}
    \caption{Top-1 accuracy (\%) of MB$^2$L across different image processing methods on the THINGS-EEG dataset.}
    \label{fig:img}
  \end{minipage}
\end{figure}

To investigate the impact of different image processing methods and validate ABVP, we applied five techniques—color jittering, grayscaling, Gaussian noise, low resolution, and mosaic—in addition to Gaussian blurring, under two conditions: with and without ABVP. Results are shown in Figure~\ref{fig:img process}.

Figure~\ref{fig:img} shows the TOP-1 accuracy of various image processing methods on the test set. Vertically, introducing ABVP improved performance for all methods except color jitter, which alters rather than weakens image content. Notably, adding Gaussian noise achieved a TOP-1 accuracy of 86.2\%, potentially reflecting human visual perception more closely. Horizontally, without ABVP, performance fluctuated around that of the originals: global weakening can degrade central, task-relevant features, while peripheral information contains redundancy, and their interplay leads to varying accuracy across methods. Detailed results and more analyses are provided in Appendix~\ref{a_img_process}.

    
       
        
We further conducted a systematic comparison of different prior forms for the adaptive blurring strategy. Beyond the logistic gating function, we evaluated parameterized priors based on exponential and quadratic functions, as well as a fully independent adaptive learning approach without explicit priors. As summarized in Table \ref{tab:visual prior}, the logistic gating function achieves the best performance, likely due to its consistency with retinal topography. Detailed results for each prior and subject are provided in Appendix~\ref{a_priors}.

\begin{wraptable}{R}{0.42\textwidth}
  \centering
  \vspace{-18pt} 
  \setlength{\tabcolsep}{10pt}
  \caption{Comparison of different visual priors in the ABVP(\%)}
  \begin{tabular}{lcc}
    \toprule
    \textbf{Visual Prior}  & \textbf{Top-1} & \textbf{Top-5} \\
    \midrule
    w/o & 72.8 & 94.8 \\
    exp & \underline{77.7} & \underline{96.6} \\
    quad & 75.6 & 95.7 \\
    \rowcolor{gray!20}
    logistic & \textbf{80.5} & \textbf{97.6} \\
    \bottomrule
  \end{tabular}
  \label{tab:visual prior}
  \vspace{-8pt}
\end{wraptable}
\vspace{0pt} 
To analyze the effect of different visual prior functions, we evaluate four types of curve functions in our adaptive blur module: without prior, exponential, quadratic, and logistic. As shown in Table~\ref{tab:visual prior}, the logistic function achieves the best performance, surpassing other curves by a clear margin. This demonstrates that the smooth and bounded characteristics of the logistic function better fit the visual prior distribution of human EEG signals.

\subsection{Channel Attention Analysis}
\begin{figure*}[ht]
  \centering
  \includegraphics[width=0.95\textwidth]{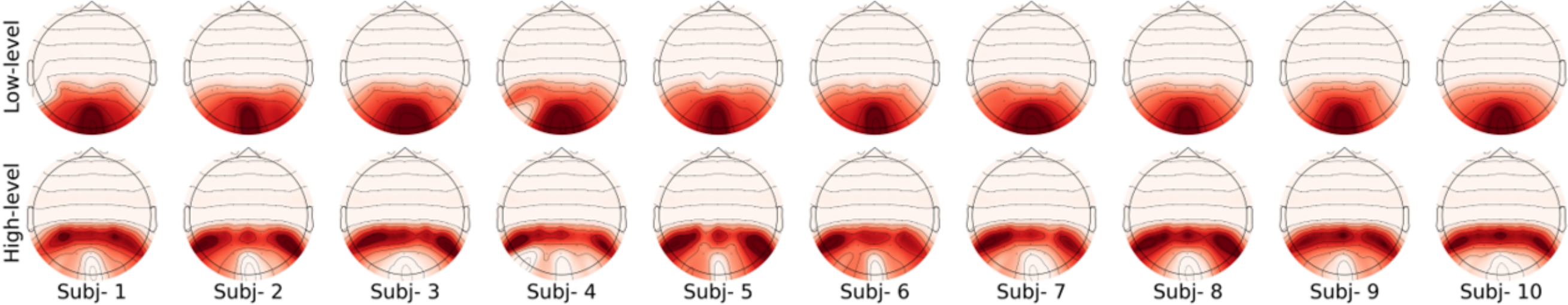}
  \caption{EEG channel attention heatmaps across subjects for low- and high-level features.}
  \label{brain_hot}
\end{figure*}

As shown in Figure~\ref{brain_hot}, we present channel attention weights after learning low- and high-level visual features. The results indicate that low-level visual features are more focused on neural responses in the occipital lobe and occipito-parietal junction, while high-level visual features are more concentrated in the parietal lobe, occipito-parietal junction, and posterior temporal lobe. This aligns with existing physiological knowledge, further validating the effectiveness of our channel prior.
\vspace{-5pt}
\begin{figure}[ht]
  \centering
  \begin{minipage}{0.48\textwidth}
    \centering
    \includegraphics[width=\linewidth]{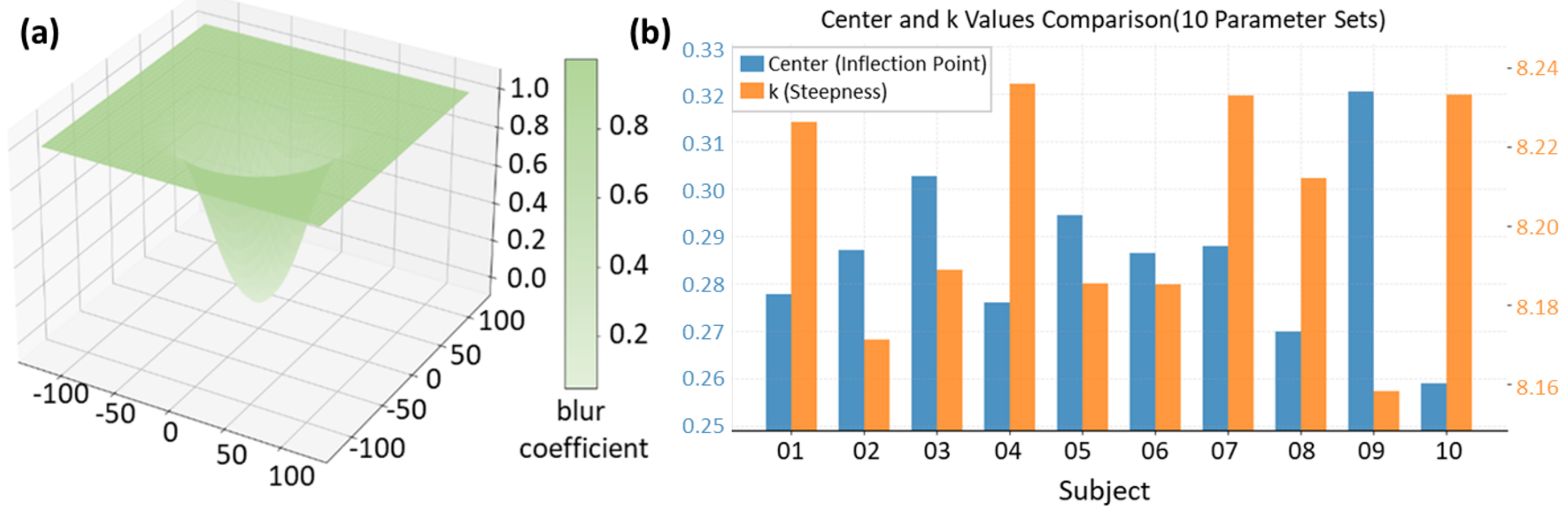}
    \caption{Visualization of adaptive blur function with visual priors: (a) Initialized blur function; (b) Learned blur function.}
    \label{blur}
  \end{minipage}
  \hfill
  \begin{minipage}{0.48\textwidth}
    \centering
    \includegraphics[width=\linewidth]{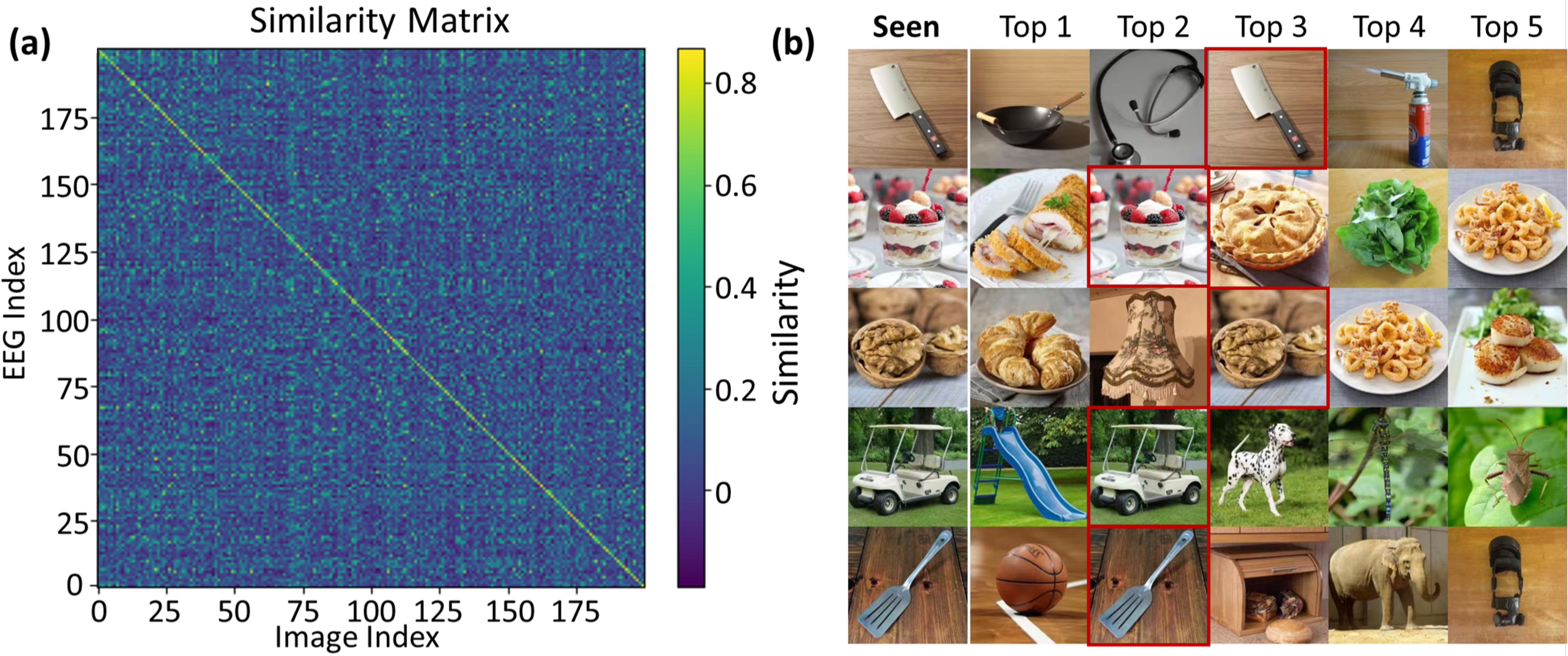}
    \caption{EEG–image alignment visualization: (a) Similarity matrix; (b) Retrieval results.}
    \label{sim}
  \end{minipage}
\end{figure}

\subsection{Visualization Analysis of Blur Function}

To investigate individual differences in visual blur, we visualize learned blur functions. As shown in Figure~\ref {blur}, even under a shared central fovea prior, participants exhibit substantial variations in spatial decay distribution across visual space, reflecting heterogeneous retinal topography and cortical organization, demonstrating the model’s ability to capture subject-specific visual characteristics.

\subsection{Semantic Alignment Analysis and Visualization}
To assess semantic alignment between EEG signals and visual stimuli, we analyze Subject 8 by computing cross-modal similarity scores for all 200 test concepts. Top five retrieved images for TOP-1 misclassified samples are shown. Figure~\ref{sim}(a) shows the similarity matrix, revealing strong cross-modal alignment, while Figure~\ref{sim}(b) indicates that, despite TOP-1 misclassifications, the top five candidates still closely resemble correct samples in semantic content and low-level features. Further results for additional test concepts and representative subjects are provided in Appendix~\ref{a_sim_all} and \ref{pictures}.
\subsection{Ablation Study on Various Encoders}

\begin{wrapfigure}{R}{0.48\textwidth} 
  \vspace{-50pt} 
  \centerline{\includegraphics[width=0.48 \columnwidth]{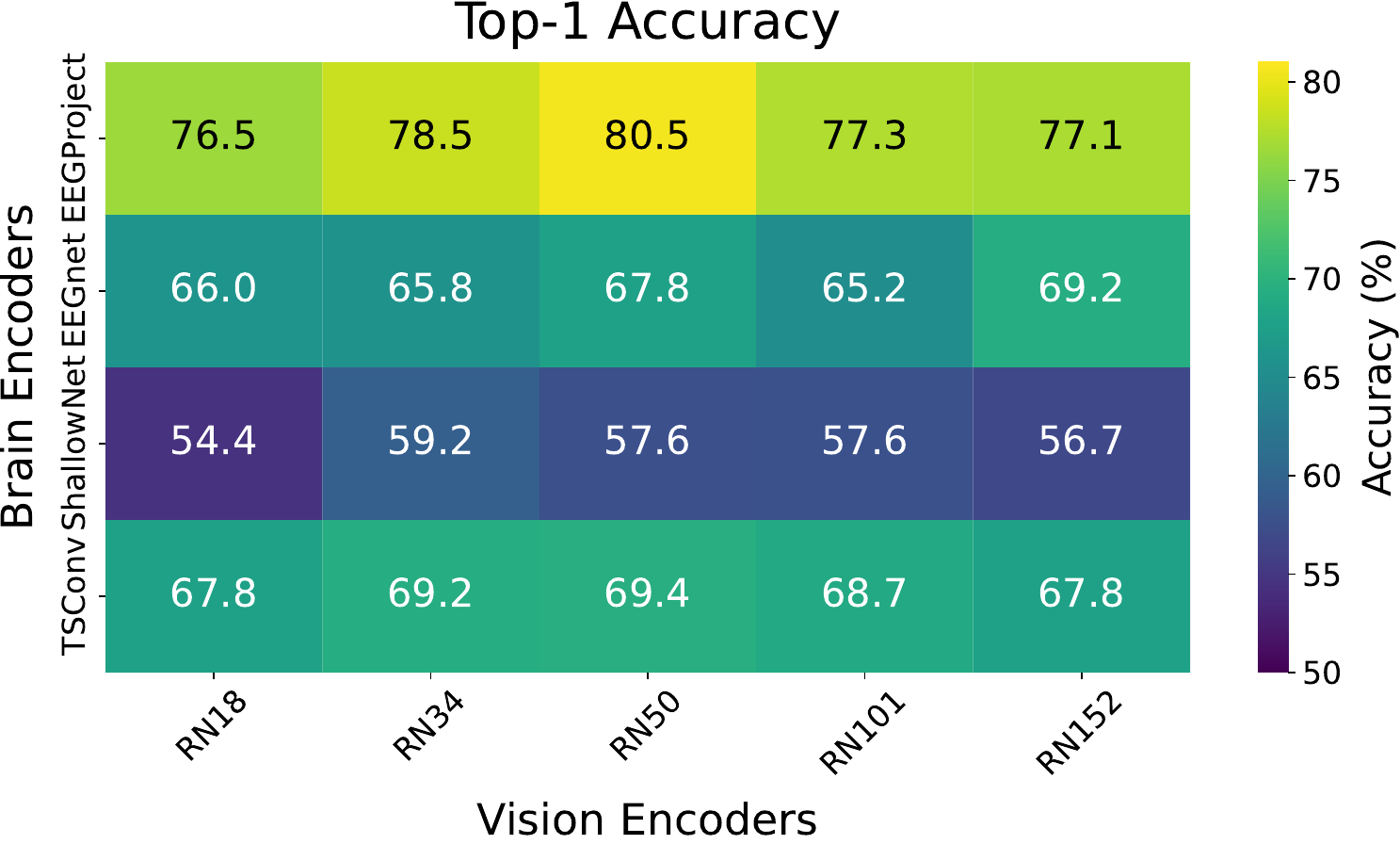}}
    \caption{Top-1 accuracy (\%) of MB$^2$L across various brain and vision encoder combinations on the THINGS-EEG.}
    \label{encoder}
  \label{fig:your_label}
\end{wrapfigure}
  \vspace{0pt}
To verify the generalizability of our framework, we conducted comprehensive experiments by training over one thousand models with four EEG encoders and five image encoders spanning diverse architectural variants, including both lightweight and deep models. Across all settings, our framework consistently outperforms comparison methods, demonstrating stable and reliable performance gains across encoder combinations, regardless of specific architectural choices. The Top-1 accuracy results on the THINGS-EEG dataset are illustrated in Figure~\ref{encoder}, while detailed quantitative results, ablation studies, and additional analyses are provided in Appendix~\ref{a_encoder}, further supporting the robustness of our approach.

\section{Conclusion}
We propose MB$^2$L, a biomimetic framework for EEG-based visual neural decoding that leverages physiologically grounded visual priors to improve cross-modal alignment. By modeling foveated perception and hierarchical visual representations, it reduces perceptual redundancy and enhances semantic consistency between EEG signals and visual stimuli. Experiments on THINGS-EEG and THINGS-MEG show consistent gains over existing methods. However, EEG temporal dynamics are modeled generically, and the method assumes a fixed central fixation point without considering microsaccades, which may limit naturalistic fidelity and suggest room for further improvement.
{
\small

\bibliographystyle{IEEEtran}
\bibliography{references}  

\newpage
\appendix
\onecolumn
\section{Experimental details}
\subsection{Datasets details}
\label{datasets}
THINGS-EEG \cite{gifford2022large} is a large-scale EEG dataset involving 10 subjects, collected using the Rapid Serial Visual Presentation (RSVP) paradigm \cite{grootswagers2019representational,intraub1981rapid,keysers2001speed}. The EEG data were recorded with a 64-channel EASYCAP system using the standard 10-10 electrode placement \cite{nuwer1998ifcn}. The training set consists of 1654 concepts, each with 10 images, with each image repeated 4 times (i.e., 1654 concepts × 10 images/concept × 4 trials/image). The test set includes 200 concepts, each with 1 image, repeated 80 times (i.e., 200 concepts × 1 image/concept × 80 trials/image). Data preprocessing follows the method detailed in \cite{song2023decoding,wu2025bridging}, where the raw EEG data is bandpass filtered between 0.1 Hz and 100 Hz, with a sampling rate of 1000 Hz. Out of the 63 channels, 17 \footnote{P7, P5, P3, P1, Pz, P2, P4, P6, P8, PO7, PO3, POz, PO4, PO8, O1, Oz, O2} electrodes located in the occipital and parietal areas related to visual processing are selected for analysis. To improve the signal-to-noise ratio (SNR), EEG repetitions are averaged, resulting in 16,540 training samples and 200 test samples per subject. The EEG data are stored in float16 format to enable faster reading speeds and reduce storage requirements.

THINGS-MEG \cite{hebart2023things} involves 4 participants and is recorded with 271 channels. The experimental design includes a stimulus duration of 500 ms, followed by a blank screen lasting 1000 ± 200 ms. The training set contains 1854 concepts × 12 images × 1 repetition, and the test set includes 200 concepts × 1 image × 12 repetitions. During data processing, 200 test concepts are removed from the training set to construct a zero-shot task. The MEG data is segmented into trials ranging from 0 to 1000 ms after the stimulus onset. Preprocessing involves applying a bandpass filter between 0.1 Hz and 100 Hz, followed by baseline correction after down-sampling the data to 200 Hz. To enhance the SNR, all MEG repetitions of an image are averaged. The data are also stored in float16 format to optimize reading speed and minimize storage usage.
\subsection{Implementation details}
\label{Imp details}
\textbf{Environment.} Our method is implemented in Python 3.10 and accelerated using PyTorch 2.0.1 with CUDA 11.8. The experimental environment is managed via Conda. Core dependencies include PyTorch, TorchVision, Transformers, Diffusers, NumPy, and SciPy. A complete list of required libraries is provided in the repository for reproducibility.All experiments are conducted on a workstation equipped with an Intel Core i7-14650HX CPU, a single NVIDIA RTX 3090 GPU with 24 GB memory, and 64 GB RAM. 

\textbf{Training Configuration}. We train the model with a batch size of 256. Except for the EEG projector, all modules are trained for 60 epochs to achieve better convergence. A unified learning rate of 1e-4 is adopted across all experiments. Model parameters are optimized using the AdamW \cite{loshchilov2017decoupled} optimizer with a weight decay of 1e-4. Early stopping is employed based on training loss and validation performance to mitigate overfitting. For the intra-subject setting, the loss coefficient of the high-level visual channel is set to 0.5. In the inter-subject setting, this coefficient is reduced to 0.1 to alleviate the impact of inter-individual variability in semantic understanding. All other hyperparameters are kept consistent across experiments.
\subsection{Comparison method}
\label{sec:comparison_method}
\begin{itemize}
  \item \textbf{BraVL}~\cite{du2023decoding} proposes a multimodal learning framework that jointly models brain signals, visual features, and linguistic representations using a mixture-of-experts mechanism. By dynamically weighting modality-specific experts, BraVL aims to enhance cross-modal alignment for EEG-based visual decoding.

  \item \textbf{NICE}~\cite{song2023decoding} is a self-supervised EEG representation learning approach that leverages contrastive objectives to learn discriminative neural features. It incorporates attention mechanisms to capture spatial dependencies across EEG channels and improves generalization under limited labeled data.

  \item \textbf{ATM}~\cite{li2024visual} introduces an adaptive EEG encoder that integrates positional encoding with spatiotemporal modeling. The method emphasizes dynamic temporal feature extraction and spatial structure awareness to enhance EEG-based visual decoding performance.

  \item \textbf{Neural-MCRL}~\cite{li2024neural} performs visual decoding by learning multimodal contrastive representations that align EEG signals with visual embeddings. It adopts a cross-modal contrastive learning strategy to reduce the semantic gap between neural and visual modalities.

  \item \textbf{CognitionCapturer}~\cite{zhang2025cognitioncapturer} focuses on modeling cognitive processes by integrating multimodal information for EEG-based visual decoding. The method aligns neural signals with visual representations through a unified semantic space to improve decoding robustness.

  \item \textbf{VE-SDN}~\cite{chen2024visual} constructs a shared semantic space via a semantic decoupling strategy, separating visual semantics from modality-specific representations. This design facilitates more effective alignment between EEG and visual features during decoding.

  \item \textbf{UBP}~\cite{wu2025bridging} introduces uncertainty-aware modeling by explicitly incorporating uncertainty priors into EEG representations. The approach enhances robustness and generalization by accounting for noise and variability inherent in neural signals.
\end{itemize}

\section{Results details}

\subsection{Ablation Study of the Proposed Framework}
\label{abl}
Table~\ref{tab:abvp_bvfe_mbcl_combinations} reports the detailed ablation results under different combinations of ABVP, BVFE, and MBCL. Removing all three components leads to severely degraded performance, indicating that effective EEG–image alignment cannot be achieved without structural or visual priors. Introducing BVFE alone does not improve performance and even results in a slight decline, suggesting that enforcing EEG-side structural constraints without addressing modality mismatch may amplify noise.

In contrast, incorporating MBCL yields a substantial performance gain, highlighting its critical role in mitigating heterogeneity between EEG signals and visual representations. When combined with MBCL, both ABVP and BVFE consistently provide additional improvements, demonstrating their complementary effects. The full model, which integrates all three components, achieves the best performance across all subjects, confirming that ABVP, BVFE, and MBCL jointly contribute to robust and effective cross-modal representation learning.
\begin{table*}[htbp]
\centering
\caption{Top-1 and Top-5 Accuracy (\%) of Different ABVP/BVFE/MBCL Combinations on EEG Dataset}
\label{tab:abvp_bvfe_mbcl_combinations}
\resizebox{\textwidth}{!}{
\begin{tabular}{ccc|cccccccccc|c}
    \toprule
    \textbf{ABVP} & \textbf{BVFE} & \textbf{MBCL} & Sub 1 & Sub 2 & Sub 3 & Sub 4 & Sub 5 & Sub 6 & Sub 7 & Sub 8 & Sub 9 & Sub 10 & Average \\
    \midrule
    \multirow{2}{*}{\xmark} & \multirow{2}{*}{\xmark} & \multirow{2}{*}{\xmark}
    & 31.00 & 24.00 & 26.00 & 30.50 & 17.00 & 28.00 & 27.50 & 32.50 & 24.50 & 30.50 & 27.15 \\
    & & & 62.00 & 58.00 & 60.50 & 60.00 & 47.50 & 61.00 & 61.00 & 63.50 & 55.50 & 64.50 & 59.35 \\
    \midrule
    \multirow{2}{*}{\xmark} & \multirow{2}{*}{\cmark} & \multirow{2}{*}{\xmark}
    & 26.00 & 28.50 & 19.50 & 29.50 & 16.00 & 28.00 & 23.50 & 31.00 & 21.50 & 29.50 & 25.30 \\
    & & & 57.00 & 54.50 & 59.00 & 55.00 & 42.00 & 60.50 & 53.00 & 59.00 & 47.50 & 57.50 & 54.50 \\
    \midrule
    \multirow{2}{*}{\xmark} & \multirow{2}{*}{\xmark} & \multirow{2}{*}{\cmark}
    & 69.00 & 70.00 & 65.50 & 57.00 & 52.50 & 74.50 & 64.00 & 68.00 & 61.50 & 72.00 & 65.40 \\
    & & & 96.50 & 93.00 & 93.00 & 87.00 & 85.50 & 95.50 & 92.00 & 96.00 & 87.00 & 95.50 & 92.10 \\
    \midrule
    \multirow{2}{*}{\cmark} & \multirow{2}{*}{\xmark} & \multirow{2}{*}{\cmark}
    & 78.50 & 79.50 & 72.50 & 65.00 & 65.50 & 83.00 & 74.50 & 81.50 & 73.00 & 85.50 & 75.85 \\
    & & & 98.00 & 98.50 & 96.50 & 94.00 & 93.50 & 98.50 & 95.50 & 99.50 & 92.50 & 99.50 & 96.60 \\
    \midrule
    \multirow{2}{*}{\xmark} & \multirow{2}{*}{\cmark} & \multirow{2}{*}{\cmark}
    & 72.50 & 71.00 & 73.00 & 59.50 & 60.50 & 76.50 & 71.00 & 72.50 & 68.50 & 71.50 & 69.65 \\
    & & & 95.00 & 96.00 & 95.50 & 89.50 & 90.00 & 95.00 & 93.00 & 97.00 & 92.00 & 97.00 & 94.00 \\
    \midrule
    \multirow{2}{*}{\cmark} & \multirow{2}{*}{\cmark} & \multirow{2}{*}{\cmark}
    &  \textbf{82.50} & \textbf{85.00} & \textbf{80.00} & \textbf{67.00} & \textbf{74.50} & \textbf{86.50} & \textbf{78.50} & \textbf{86.50} & \textbf{77.50} & \textbf{86.50} & \textbf{80.45} \\
     & & &  \textbf{99.00} & \textbf{99.00} & \textbf{98.00} & \textbf{95.00} & \textbf{99.50} & \textbf{98.50} & \textbf{96.00} & \textbf{99.50} & \textbf{95.50} & \textbf{99.50} & \textbf{97.95} \\
    \bottomrule
\end{tabular}
}
\end{table*}
\subsection{Results of Visual and EEG Encoder Variants}
\label{a_encoder}
Tables~\ref{tab:brain_encoders} and~\ref{tab:vision_encoders} summarize the architectural configurations of the brain and vision encoders used in our experiments. The compared models span a wide range of parameter scales and embedding dimensions, enabling a fair and comprehensive analysis of architectural effects.

For the brain encoders, we include both lightweight CNN-based architectures (e.g., EEGNet, Shallownet, and TSConv) and a higher-capacity projection-based encoder, covering commonly adopted designs for EEG representation learning. For the vision encoders, multiple ResNet variants with increasing depth and model capacity are evaluated, allowing us to examine the influence of visual encoder scale and embedding dimensionality on cross-modal alignment. This design ensures that the observed performance differences can be attributed to architectural characteristics rather than model size bias.

\begin{table*}[htbp]
  \centering
  \setlength{\tabcolsep}{8pt}
  \begin{minipage}[t]{0.48\textwidth}
    \centering
    \caption{Details of Brain Encoders}
    \label{tab:brain_encoders}
    \begin{tabular}{cc}
      \toprule
      \textbf{Brain Encoder} & \textbf{Params} \\
      \midrule
      EEGProjectLayer & 5.40 M \\
      EEGNet & 2.34 M \\
      Shallownet & 2.56 M \\
      TSconv & 2.56 M \\
      \bottomrule
    \end{tabular}
  \end{minipage}
  \hfill
  \begin{minipage}[t]{0.48\textwidth}
    \centering
    \caption{Details of Vision Encoders}
    \label{tab:vision_encoders}
    \begin{tabular}{ccc}
      \toprule
      \textbf{Vision Encoder} & \textbf{Params} & \textbf{Emb dim} \\
      \midrule
      ResNet18 & 11.69 M & 512 \\
      ResNet34 & 21.80 M & 512 \\
      ResNet50 & 38.32 M & 2048 \\
      ResNet101 & 56.26 M & 2048 \\
      ResNet152 & 60.19 M & 2048 \\
      \bottomrule
    \end{tabular}
  \end{minipage}
\end{table*}

To further analyze the generalization behavior of the proposed framework, we report performance comparisons across different combinations of EEG encoders and ResNet-based image encoders with varying depths, as summarized in Tables~\ref{ResNet18-based Model}–\ref{tab:resnet152_based_model}. These experiments involve four representative EEG encoders and five visual backbones, resulting in over one thousand trained model variants.

Across all visual backbones, EEGProjectLayer consistently achieves the best Top-1 and Top-5 performance, indicating its strong ability to learn stable and discriminative EEG representations. In contrast, EEGNet, ShallowNet, and TSConv show larger performance fluctuations across subjects, especially when paired with deeper image encoders, suggesting reduced robustness to inter-subject variability.

Regarding visual backbone depth, performance improves from ResNet18 to ResNet50 but does not consistently benefit from deeper architectures such as ResNet101 and ResNet152. This indicates that increasing image encoder capacity alone is insufficient for improving EEG-based visual decoding and may even hinder optimization.Overall, these results demonstrate that the framework exhibits stable performance trends across different encoder configurations.
\begin{table*}[htbp]
  \centering
  \caption{Top-1 and Top-5 Accuracy (\%) of Different ResNet18-based Model Architectures on EEG Dataset}
  \label{ResNet18-based Model}
  \resizebox{\textwidth}{!}{
  \begin{tabular}{lc|cccccccccc|c}
    \toprule
    Method & & Sub 1 & Sub 2 & Sub 3 & Sub 4 & Sub 5 & Sub 6 & Sub 7 & Sub 8 & Sub 9 & Sub 10 & Average \\
    \midrule
    \multirow{2}{*}{EEGProjectLayer}
    & Top-1 & 76.00 & 83.50 & 76.50 & 63.50 & 72.00 & 81.50 & 76.00 & 83.50 & 73.00 & 79.00 & 76.45 \\
    & Top-5 & 96.00 & 99.00 & 97.50 & 91.00 & 93.50 & 97.50 & 94.50 & 99.00 & 92.50 & 98.00 & 95.85 \\
    \midrule
    \multirow{2}{*}{EEGNet}
    & Top-1 & 71.50 & 68.50 & 59.00 & 52.00 & 66.50 & 73.00 & 60.00 & 76.00 & 61.50 & 72.00 & 66.00 \\
    & Top-5 & 91.50 & 96.50 & 90.00 & 84.00 & 93.50 & 96.50 & 88.00 & 95.00 & 89.00 & 95.00 & 91.90 \\
    \midrule
    \multirow{2}{*}{Shallownet}
    & Top-1 & 58.50 & 55.50 & 56.00 & 51.50 & 42.50 & 63.50 & 51.50 & 57.50 & 47.50 & 60.00 & 54.40 \\
    & Top-5 & 90.00 & 89.50 & 89.00 & 81.50 & 82.50 & 90.50 & 84.00 & 89.00 & 81.00 & 88.00 & 86.50 \\
    \midrule
    \multirow{2}{*}{TSconv}
    & Top-1 & 70.50 & 69.00 & 67.00 & 55.50 & 63.00 & 77.50 & 66.50 & 77.00 & 61.00 & 71.00 & 67.80 \\
    & Top-5 & 96.00 & 93.00 & 95.50 & 84.00 & 93.00 & 96.00 & 88.50 & 95.50 & 87.00 & 95.00 & 92.35 \\
    \bottomrule
  \end{tabular}
  }
\end{table*}
\begin{table*}[htbp]
  \centering
  \caption{Top-1 and Top-5 Accuracy (\%) of Different ResNet34-based Model Architectures on EEG Dataset}
  \label{tab:resnet34_based_model}
  \resizebox{\textwidth}{!}{
  \begin{tabular}{lc|cccccccccc|c}
    \toprule
    Method & & Sub 1 & Sub 2 & Sub 3 & Sub 4 & Sub 5 & Sub 6 & Sub 7 & Sub 8 & Sub 9 & Sub 10 & Average \\
    \midrule
    \multirow{2}{*}{EEGProjectLayer}
    & Top-1 & 78.00 & 79.00 & 81.00 & 67.50 & 72.50 & 86.00 & 80.00 & 85.50 & 74.50 & 80.50 & 78.45 \\
    & Top-5 & 97.50 & 99.00 & 99.50 & 94.00 & 95.00 & 98.00 & 96.00 & 99.50 & 94.50 & 99.00 & 97.20 \\
    \midrule
    \multirow{2}{*}{EEGNet}
    & Top-1 & 75.00 & 66.00 & 61.00 & 51.00 & 58.50 & 76.50 & 58.00 & 76.50 & 62.00 & 73.50 & 65.80 \\
    & Top-5 & 94.50 & 96.00 & 91.50 & 84.50 & 88.00 & 96.50 & 87.00 & 94.50 & 91.50 & 96.00 & 92.00 \\
    \midrule
    \multirow{2}{*}{Shallownet}
    & Top-1 & 64.00 & 60.00 & 63.50 & 52.00 & 55.00 & 72.00 & 63.00 & 61.50 & 45.00 & 56.50 & 59.25 \\
    & Top-5 & 91.50 & 91.00 & 94.00 & 84.50 & 87.00 & 95.50 & 88.00 & 91.50 & 76.50 & 88.50 & 88.80 \\
    \midrule
    \multirow{2}{*}{TSconv}
    & Top-1 & 74.00 & 68.50 & 71.50 & 52.00 & 63.50 & 82.00 & 65.50 & 78.50 & 62.00 & 74.50 & 69.20 \\
    & Top-5 & 95.50 & 94.00 & 95.00 & 88.00 & 90.50 & 96.50 & 89.00 & 95.50 & 89.00 & 97.00 & 93.00 \\
    \bottomrule
  \end{tabular}
  }
\end{table*}

\begin{table*}[htbp]
  \centering
  \caption{Top-1 and Top-5 Accuracy (\%) of Different ResNet50-based Model Architectures on EEG Dataset}
  \label{tab:resnet50_based_model}
  \resizebox{\textwidth}{!}{
  \begin{tabular}{lc|cccccccccc|c}
    \toprule
    Method & & Sub 1 & Sub 2 & Sub 3 & Sub 4 & Sub 5 & Sub 6 & Sub 7 & Sub 8 & Sub 9 & Sub 10 & Average \\
    \midrule
    \multirow{2}{*}{EEGProjectLayer}
    & Top-1 & 82.50 & 85.00 & 80.00 & 67.00 & 74.50 & 86.50 & 78.50 & 86.50 & 77.50 & 86.50 & 80.45 \\
    & Top-5 & 99.00 & 99.00 & 98.00 & 95.00 & 99.50 & 98.50 & 96.00 & 99.50 & 95.50 & 99.50 & 97.95 \\
    \midrule
    \multirow{2}{*}{EEGNet}
    & Top-1 & 73.50 & 69.00 & 60.00 & 55.00 & 63.00 & 81.50 & 60.50 & 76.00 & 62.00 & 77.50 & 67.80 \\
    & Top-5 & 94.00 & 94.50 & 92.00 & 87.00 & 95.50 & 98.00 & 90.50 & 95.50 & 91.50 & 96.00 & 93.45 \\
    \midrule
    \multirow{2}{*}{Shallownet}
    & Top-1 & 71.00 & 68.50 & 51.00 & 49.50 & 42.50 & 72.00 & 56.50 & 60.50 & 49.50 & 55.00 & 57.60 \\
    & Top-5 & 93.00 & 92.00 & 87.00 & 82.50 & 79.00 & 96.50 & 85.50 & 91.00 & 81.50 & 86.50 & 87.45 \\
    \midrule
    \multirow{2}{*}{TSconv}
    & Top-1 & 74.00 & 66.50 & 66.50 & 54.00 & 64.00 & 85.00 & 64.50 & 76.50 & 66.00 & 77.00 & 69.40 \\
    & Top-5 & 95.00 & 93.00 & 94.50 & 83.00 & 91.00 & 98.50 & 89.00 & 96.50 & 93.00 & 97.00 & 93.05 \\
    \bottomrule
  \end{tabular}
  }
\end{table*}

\newpage
\begin{table*}[htbp]
  \centering
  \caption{Top-1 and Top-5 Accuracy (\%) of Different ResNet101-based Model Architectures on EEG Dataset}
  \label{tab:resnet101_based_model}
  \resizebox{\textwidth}{!}{
  \begin{tabular}{lc|cccccccccc|c}
    \toprule
    Method & & Sub 1 & Sub 2 & Sub 3 & Sub 4 & Sub 5 & Sub 6 & Sub 7 & Sub 8 & Sub 9 & Sub 10 & Average \\
    \midrule
    \multirow{2}{*}{EEGProjectLayer}
    & Top-1 & 81.50 & 81.50 & 78.00 & 63.50 & 70.50 & 81.00 & 76.00 & 84.50 & 72.50 & 84.00 & 77.30 \\
    & Top-5 & 98.50 & 97.50 & 97.00 & 91.00 & 94.00 & 97.50 & 94.00 & 99.50 & 93.00 & 98.50 & 96.05 \\
    \midrule
    \multirow{2}{*}{EEGNet}
    & Top-1 & 71.00 & 69.00 & 57.50 & 50.50 & 57.50 & 78.00 & 61.50 & 75.50 & 60.00 & 71.00 & 65.15 \\
    & Top-5 & 92.00 & 94.00 & 89.00 & 81.50 & 87.00 & 98.50 & 86.00 & 95.50 & 89.00 & 95.50 & 90.80 \\
    \midrule
    \multirow{2}{*}{Shallownet}
    & Top-1 & 61.00 & 65.00 & 61.50 & 48.00 & 53.00 & 70.00 & 58.00 & 64.00 & 47.00 & 49.00 & 57.65 \\
    & Top-5 & 89.50 & 92.00 & 93.00 & 85.50 & 86.00 & 95.00 & 87.50 & 91.50 & 84.00 & 85.50 & 88.95 \\
    \midrule
    \multirow{2}{*}{TSconv}
    & Top-1 & 73.00 & 73.00 & 67.50 & 52.00 & 64.50 & 79.50 & 64.50 & 82.50 & 60.00 & 70.50 & 68.70 \\
    & Top-5 & 96.50 & 94.00 & 93.00 & 85.50 & 91.00 & 97.00 & 89.00 & 98.50 & 89.00 & 96.50 & 93.00 \\
    \bottomrule
  \end{tabular}
  }
\end{table*}

\begin{table*}[htbp]
  \centering
  \caption{Top-1 and Top-5 Accuracy (\%) of Different ResNet152-based Model Architectures on EEG Dataset}
  \label{tab:resnet152_based_model}
  \resizebox{\textwidth}{!}{
  \begin{tabular}{lc|cccccccccc|c}
    \toprule
    Method & & Sub 1 & Sub 2 & Sub 3 & Sub 4 & Sub 5 & Sub 6 & Sub 7 & Sub 8 & Sub 9 & Sub 10 & Average \\
    \midrule
    \multirow{2}{*}{EEGProjectLayer}
    & Top-1 & 79.00 & 81.00 & 78.00 & 65.00 & 71.00 & 82.00 & 75.00 & 82.00 & 77.00 & 81.00 & 77.10 \\
    & Top-5 & 96.50 & 97.00 & 96.50 & 92.50 & 93.00 & 97.00 & 95.50 & 99.50 & 93.50 & 98.00 & 95.90 \\
    \midrule
    \multirow{2}{*}{EEGNet}
    & Top-1 & 76.50 & 72.50 & 62.00 & 54.00 & 65.00 & 79.00 & 64.50 & 77.50 & 66.00 & 74.50 & 69.15 \\
    & Top-5 & 94.50 & 93.50 & 89.50 & 84.00 & 92.50 & 97.00 & 90.00 & 96.00 & 92.00 & 97.00 & 92.60 \\
    \midrule
    \multirow{2}{*}{Shallownet}
    & Top-1 & 59.00 & 51.20 & 50.00 & 57.00 & 59.00 & 74.00 & 69.00 & 62.00 & 40.50 & 45.50 & 56.72 \\
    & Top-5 & 88.00 & 85.00 & 86.50 & 87.00 & 89.00 & 94.50 & 87.50 & 93.00 & 72.00 & 81.50 & 86.40 \\
    \midrule
    \multirow{2}{*}{TSconv}
    & Top-1 & 75.50 & 66.50 & 66.50 & 56.00 & 62.00 & 79.50 & 62.00 & 77.00 & 60.50 & 72.00 & 67.75 \\
    & Top-5 & 97.00 & 93.00 & 93.00 & 86.50 & 92.50 & 97.50 & 88.00 & 95.00 & 88.50 & 95.00 & 92.60 \\
    \bottomrule
  \end{tabular}
  }
\end{table*}
\subsection{Results under Different Image Processing Methods}
\label{a_img_process}

\begin{figure}[htbp]
  \begin{center}
    \centerline{\includegraphics[width=\textwidth]{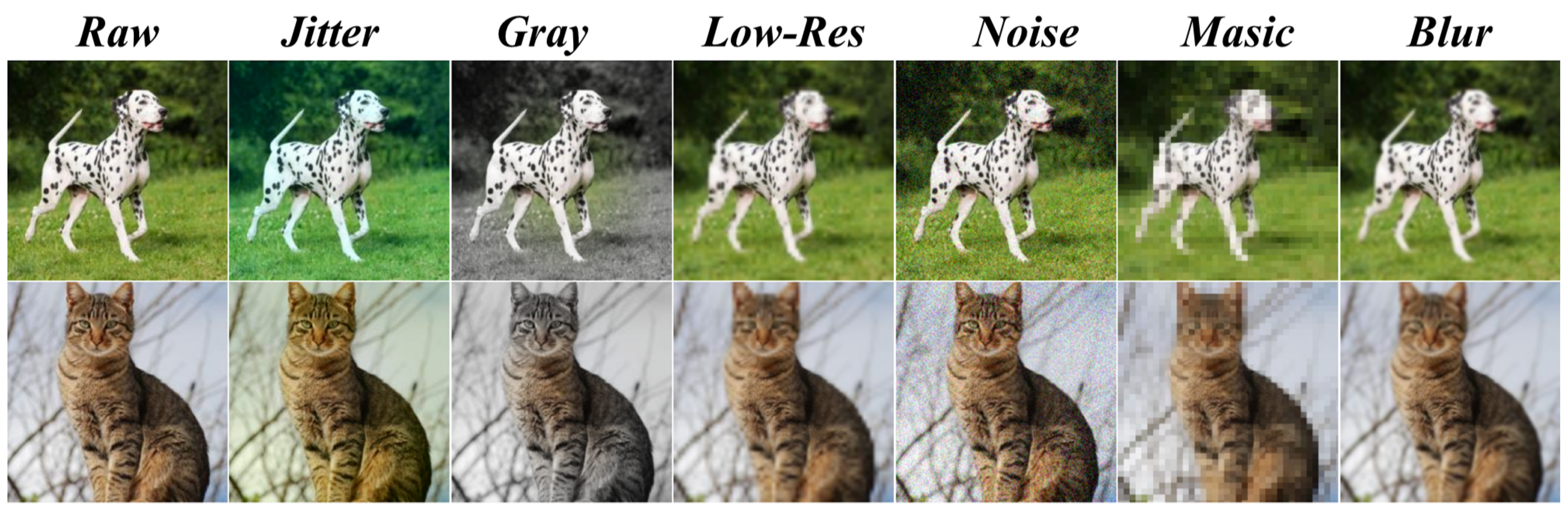}}
    \caption{Effect of Incorporating ABVP across Different Image Processing Methods}
    \label{fig:img2}
  \end{center}
\end{figure}
Tables~\ref{tab:top1_accuracy} and~\ref{tab:top5_accuracy} report the Top-1 and Top-5 accuracy of different image processing methods with and without ABVP, while Figure~\ref{fig:img2} qualitatively illustrates how ABVP modulates visual degradation across different processing strategies. Overall, introducing ABVP consistently improves performance across most image processing methods, demonstrating the robustness and general applicability of the proposed visual prior. Notable gains are observed for degradation-based methods such as Gaussian noise, low resolution, mosaic, grayscale, and Gaussian blur, with Gaussian noise achieving the highest Top-1 accuracy when combined with ABVP. In contrast, color jitter shows only marginal improvement, as it primarily alters chromatic distributions rather than structural visual content, limiting the effectiveness of adaptive blurring.

Without ABVP, performance varies across image processing methods and remains close to that of the original images, suggesting that uniform global degradation may suppress task-relevant central information while reducing redundant peripheral cues. By explicitly modeling the foveated structure of human vision, ABVP adaptively reallocates visual information, as visually reflected in Figure~\ref{fig:img2} and quantitatively validated by the consistent improvements in both Top-1 and Top-5 accuracy.
\begin{table}[!htbp]
  \centering
  \caption{Top-1 Accuracy (\%) of Different Augmentation Methods on EEG Dataset}
  \label{tab:top1_accuracy}
  \resizebox{\textwidth}{!}{
  \begin{tabular}{lc|cccccccccc|c}
    \toprule
    Method & & Sub 1 & Sub 2 & Sub 3 & Sub 4 & Sub 5 & Sub 6 & Sub 7 & Sub 8 & Sub 9 & Sub 10 & Average \\
    \midrule
    \multirow{2}{*}{Color Jitter}
    & w/o ABVP & 73.50 & 73.00 & 71.50 & 64.00 & 61.00 & 74.50 & 72.00 & 72.00 & 67.00 & 76.00 & 70.45 \\
    & w/ ABVP & \cellcolor{gray!20}75.50 & \cellcolor{gray!20}73.50 & \cellcolor{gray!20}76.00 & \cellcolor{gray!20}62.00 & \cellcolor{gray!20}63.50 & \cellcolor{gray!20}73.00 & \cellcolor{gray!20}71.50 & \cellcolor{gray!20}76.50 & \cellcolor{gray!20}69.00 & \cellcolor{gray!20}76.00 & \cellcolor{gray!20}\textbf{71.65} \\
    \midrule
    \multirow{2}{*}{Gaussian Noise}
    & w/o ABVP & 71.50 & 71.00 & 71.50 & 57.00 & 59.50 & 73.50 & 71.00 & 73.50 & 66.50 & 76.50 & 69.15 \\
    & w/ ABVP & \cellcolor{gray!20}88.50 & \cellcolor{gray!20}87.00 & \cellcolor{gray!20}84.50 & \cellcolor{gray!20}77.00 & \cellcolor{gray!20}82.00 & \cellcolor{gray!20}90.50 & \cellcolor{gray!20}88.00 & \cellcolor{gray!20}92.50 & \cellcolor{gray!20}80.50 & \cellcolor{gray!20}91.00 & \cellcolor{gray!20}\textbf{86.15} \\
    \midrule
    \multirow{2}{*}{Low Resolution}
    & w/o ABVP & 76.00 & 73.50 & 71.50 & 59.00 & 61.50 & 76.00 & 69.00 & 76.00 & 70.00 & 75.50 & 70.80 \\
    & w/ ABVP & \cellcolor{gray!20}81.00 & \cellcolor{gray!20}84.50 & \cellcolor{gray!20}82.50 & \cellcolor{gray!20}67.00 & \cellcolor{gray!20}70.00 & \cellcolor{gray!20}85.50 & \cellcolor{gray!20}78.50 & \cellcolor{gray!20}86.00 & \cellcolor{gray!20}77.50 & \cellcolor{gray!20}86.00 & \cellcolor{gray!20}\textbf{79.85} \\
    \midrule
    \multirow{2}{*}{Mosaic}
    & w/o ABVP & 71.00 & 71.00 & 65.00 & 54.50 & 55.00 & 72.50 & 68.00 & 72.00 & 64.50 & 71.50 & 66.50 \\
    & w/ ABVP & \cellcolor{gray!20}83.00 & \cellcolor{gray!20}83.50 & \cellcolor{gray!20}81.50 & \cellcolor{gray!20}67.00 & \cellcolor{gray!20}72.00 & \cellcolor{gray!20}84.00 & \cellcolor{gray!20}78.50 & \cellcolor{gray!20}87.00 & \cellcolor{gray!20}77.50 & \cellcolor{gray!20}82.50 & \cellcolor{gray!20}\textbf{79.65} \\
    \midrule
    \multirow{2}{*}{Gray}
    & w/o ABVP & 64.00 & 66.50 & 63.50 & 48.50 & 54.50 & 64.00 & 61.00 & 63.50 & 58.00 & 68.50 & 61.20 \\
    & w/ ABVP & \cellcolor{gray!20}79.00 & \cellcolor{gray!20}83.00 & \cellcolor{gray!20}77.00 & \cellcolor{gray!20}65.50 & \cellcolor{gray!20}68.50 & \cellcolor{gray!20}79.50 & \cellcolor{gray!20}74.00 & \cellcolor{gray!20}83.00 & \cellcolor{gray!20}72.00 & \cellcolor{gray!20}81.50 & \cellcolor{gray!20}\textbf{76.30} \\
    \midrule
    \multirow{2}{*}{Gaussian blur}
    & w/o ABVP & 77.00 & 74.50 & 76.00 & 59.50 & 63.50 & 79.00 & 73.00 & 75.50 & 72.00 & 77.00 & 72.70 \\
    & w/ ABVP & \cellcolor{gray!20}82.50 & \cellcolor{gray!20}85.00 & \cellcolor{gray!20}80.00 & \cellcolor{gray!20}67.00 & \cellcolor{gray!20}74.50 & \cellcolor{gray!20}86.50 & \cellcolor{gray!20}78.50 & \cellcolor{gray!20}86.50 & \cellcolor{gray!20}77.50 & \cellcolor{gray!20}86.50 & \cellcolor{gray!20}\textbf{80.45} \\
    \bottomrule
  \end{tabular}
  }
\end{table}
\begin{table}[!htbp]
  \centering
  \caption{Top-5 Accuracy (\%) of Different Augmentation Methods on EEG Dataset}
  \label{tab:top5_accuracy}
  \resizebox{\textwidth}{!}{
  \begin{tabular}{lc|cccccccccc|c}
    \toprule
    Method & & Sub 1 & Sub 2 & Sub 3 & Sub 4 & Sub 5 & Sub 6 & Sub 7 & Sub 8 & Sub 9 & Sub 10 & Average \\
    \midrule
    \multirow{2}{*}{Color Jitter}
    & w/o ABVP & 95.50 & 96.00 & 94.00 & 90.00 & 90.00 & 96.00 & 93.50 & 97.00 & 92.50 & 98.00 & 94.25 \\
    & w/ ABVP & \cellcolor{gray!20}96.00 & \cellcolor{gray!20}95.50 & \cellcolor{gray!20}96.00 & \cellcolor{gray!20}91.00 & \cellcolor{gray!20}92.00 & \cellcolor{gray!20}95.50 & \cellcolor{gray!20}93.50 & \cellcolor{gray!20}97.50 & \cellcolor{gray!20}92.00 & \cellcolor{gray!20}98.00 & \cellcolor{gray!20}\textbf{94.70} \\
    \midrule
    \multirow{2}{*}{Gaussian Noise}
    & w/o ABVP & 95.50 & 95.50 & 93.00 & 87.50 & 90.00 & 95.00 & 93.00 & 95.50 & 91.00 & 97.00 & 93.30 \\
    & w/ ABVP & \cellcolor{gray!20}99.50 & \cellcolor{gray!20}99.00 & \cellcolor{gray!20}99.00 & \cellcolor{gray!20}97.00 & \cellcolor{gray!20}97.50 & \cellcolor{gray!20}99.50 & \cellcolor{gray!20}98.00 & \cellcolor{gray!20}100.00 & \cellcolor{gray!20}95.50 & \cellcolor{gray!20}99.50 & \cellcolor{gray!20}\textbf{98.45} \\
    \midrule
    \multirow{2}{*}{Low Resolution}
    & w/o ABVP & 98.00 & 97.00 & 95.00 & 90.00 & 90.00 & 95.00 & 63.50 & 97.00 & 93.50 & 97.00 & 91.60 \\
    & w/ ABVP & \cellcolor{gray!20}99.00 & \cellcolor{gray!20}98.50 & \cellcolor{gray!20}97.50 & \cellcolor{gray!20}94.00 & \cellcolor{gray!20}95.00 & \cellcolor{gray!20}97.50 & \cellcolor{gray!20}97.50 & \cellcolor{gray!20}99.00 & \cellcolor{gray!20}94.50 & \cellcolor{gray!20}99.50 & \cellcolor{gray!20}\textbf{97.20} \\
    \midrule
    \multirow{2}{*}{Mosaic}
    & w/o ABVP & 95.00 & 91.50 & 93.00 & 85.50 & 86.00 & 91.50 & 92.50 & 95.00 & 90.50 & 95.00 & 91.55 \\
    & w/ ABVP & \cellcolor{gray!20}99.00 & \cellcolor{gray!20}98.00 & \cellcolor{gray!20}98.90 & \cellcolor{gray!20}93.00 & \cellcolor{gray!20}95.00 & \cellcolor{gray!20}99.00 & \cellcolor{gray!20}98.50 & \cellcolor{gray!20}99.50 & \cellcolor{gray!20}95.50 & \cellcolor{gray!20}99.50 & \cellcolor{gray!20}\textbf{97.59} \\
    \midrule
    \multirow{2}{*}{Gray}
    & w/o ABVP & 90.50 & 92.50 & 90.00 & 84.50 & 84.50 & 91.50 & 90.50 & 91.50 & 90.00 & 92.50 & 89.80 \\
    & w/ ABVP & \cellcolor{gray!20}97.50 & \cellcolor{gray!20}98.00 & \cellcolor{gray!20}97.00 & \cellcolor{gray!20}97.00 & \cellcolor{gray!20}64.50 & \cellcolor{gray!20}98.00 & \cellcolor{gray!20}97.00 & \cellcolor{gray!20}99.00 & \cellcolor{gray!20}95.00 & \cellcolor{gray!20}98.00 & \cellcolor{gray!20}\textbf{94.10} \\
    \midrule
    \multirow{2}{*}{Gaussian blur}
    & w/o ABVP & 96.00 & 98.00 & 95.50 & 89.00 & 90.50 & 95.00 & 94.00 & 97.50 & 94.00 & 97.50 & 94.70 \\
    & w/ ABVP & \cellcolor{gray!20}99.00 & \cellcolor{gray!20}99.00 & \cellcolor{gray!20}98.00 & \cellcolor{gray!20}95.00 & \cellcolor{gray!20}99.50 & \cellcolor{gray!20}98.50 & \cellcolor{gray!20}96.00 & \cellcolor{gray!20}99.50 & \cellcolor{gray!20}95.50 & \cellcolor{gray!20}99.50 & \cellcolor{gray!20}\textbf{97.95} \\
    \bottomrule
  \end{tabular}
  }
\end{table}
\\

\subsection{Results under Different Visual Function Priors}
\label{a_priors}
\begin{figure}[H]
  \vskip 0.2in
  \begin{center}
    \centerline{\includegraphics[width=\textwidth]{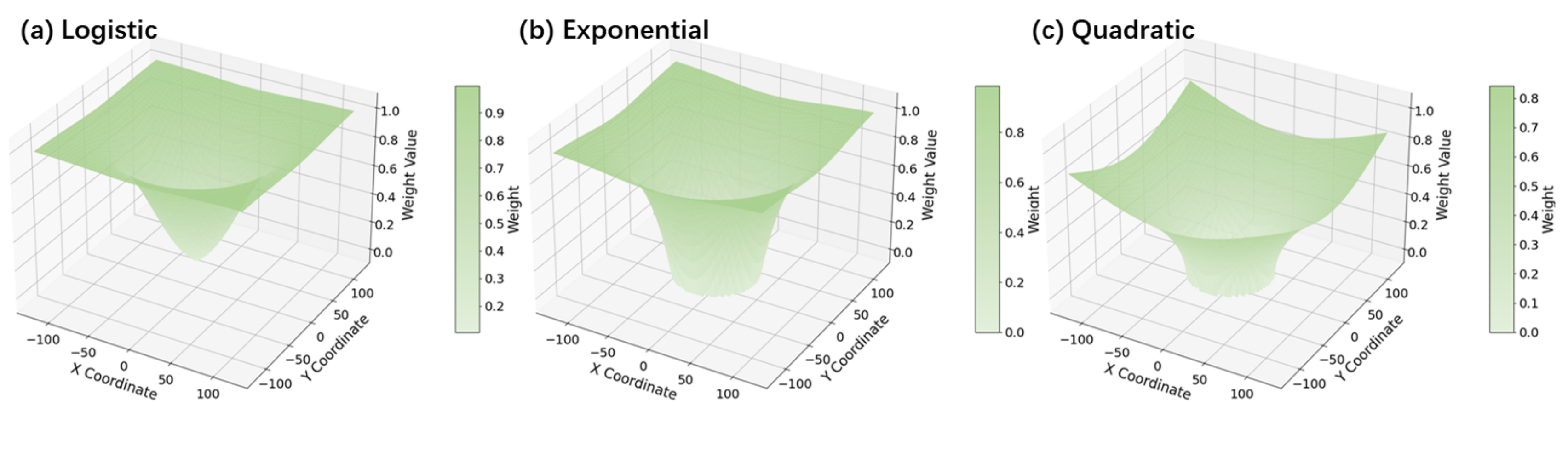}}
    \caption{Visualization of Retinal Topology Fitting Function}
    \label{fig:visual_priors_topo}
  \end{center}
\end{figure}
To investigate the impact of visual function priors on EEG-image alignment, we evaluated three biologically inspired formulations—Logistic, Exponential (Exp), and Quadratic (Quad)—alongside a baseline without priors. Figure~\ref{fig:visual_priors_topo} qualitatively illustrates their spatial modulation across the foveated visual field. The Logistic prior (Figure~\ref{fig:visual_priors_topo}a) emphasizes the central region with smoothly decaying peripheral contributions, the Exponential prior (Figure~\ref{fig:visual_priors_topo}b) increases weights gradually beyond the fovea, and the Quadratic prior (Figure~\ref{fig:visual_priors_topo}c) highlights peripheral regions with a shallower central dip.

Mathematically, the Exponential and Quadratic priors are defined as

$$w_{exp}(r) = \mathrm{clip}\big(1 - \exp(-\lambda \cdot \max(r - r_0, 0)), 0, 1\big)$$
$$w_{quad}(r) = \mathrm{clip}\big((\max(r - r_0, 0))^\gamma, 0, 1\big)$$

where $r$ is the radial distance from the fovea, $r_0$ the foveal boundary, and $\lambda$, $\gamma$ control the attenuation shape. 

Table~\ref{tab:blur_adaptive_methods} summarizes the Top-1 and Top-5 accuracy for each subject and the overall average performance. It is evident that all prior-based methods outperform the baseline without a prior, highlighting the effectiveness of incorporating structural constraints into adaptive blurring. Among the variants, the logistic prior consistently achieves the highest accuracy across most subjects, with an average Top-1 of 80.45\% and Top-5 of 97.95\%. The exponential and quadratic priors also improve performance relative to the baseline but show slightly lower Top-1 and Top-5 values, likely because their attention allocation does not perfectly match the foveated structure of human retinal topography. The fully independent variant (w/o prior) exhibits competitive results but lacks a systematic central-peripheral bias, leading to a small but noticeable performance drop.

\begin{table*}[h]
  \centering
  \caption{Top-1 and Top-5 Accuracy (\%) of Different Blur and Adaptive Methods on EEG Dataset}
  \label{tab:blur_adaptive_methods}
  \resizebox{\textwidth}{!}{
  \begin{tabular}{lc|cccccccccc|c}
    \toprule
    Method & & Sub 1 & Sub 2 & Sub 3 & Sub 4 & Sub 5 & Sub 6 & Sub 7 & Sub 8 & Sub 9 & Sub 10 & Average \\
    \midrule
    \multirow{2}{*}{Logistic}
    & Top-1 & \cellcolor{gray!20}82.50 & \cellcolor{gray!20}85.00 & \cellcolor{gray!20}80.00 & \cellcolor{gray!20}67.00 & \cellcolor{gray!20}75.00 & \cellcolor{gray!20}86.50 & \cellcolor{gray!20}78.50 & \cellcolor{gray!20}86.50 & \cellcolor{gray!20}77.50 & \cellcolor{gray!20}86.50 & \cellcolor{gray!20}80.45 \\
    & Top-5 & \cellcolor{gray!20}99.00 & \cellcolor{gray!20}99.00 & \cellcolor{gray!20}98.00 & \cellcolor{gray!20}95.00 & \cellcolor{gray!20}95.50 & \cellcolor{gray!20}98.50 & \cellcolor{gray!20}96.00 & \cellcolor{gray!20}99.50 & \cellcolor{gray!20}95.50 & \cellcolor{gray!20}99.50 & \cellcolor{gray!20}97.95 \\
    \midrule
    \multirow{2}{*}{Exp}
    & Top-1 & 80.50 & 80.50 & 76.50 & 67.50 & 70.00 & 86.00 & 77.00 & 82.00 & 74.00 & 83.00 & 77.70 \\
    & Top-5 & 97.50 & 97.50 & 97.50 & 94.00 & 95.00 & 98.00 & 95.00 & 99.00 & 94.50 & 98.50 & 96.65 \\
    \midrule
    \multirow{2}{*}{Quad}
    & Top-1 & 78.50 & 78.50 & 74.50 & 61.50 & 70.00 & 82.00 & 75.50 & 79.00 & 76.50 & 80.50 & 75.60 \\
    & Top-5 & 96.50 & 97.50 & 96.50 & 92.00 & 94.00 & 97.50 & 94.00 & 98.00 & 93.50 & 98.00 & 95.70 \\
    \midrule
    \multirow{2}{*}{w/o Prior}
    & Top-1 & 75.50 & 74.50 & 73.50 & 60.50 & 66.00 & 78.00 & 72.00 & 78.00 & 73.50 & 77.00 & 72.80 \\
    & Top-5 & 97.50 & 96.00 & 96.00 & 89.00 & 92.00 & 95.50 & 93.50 & 98.00 & 92.50 & 98.00 & 94.75 \\
    \bottomrule
  \end{tabular}
  }
\end{table*}
\newpage

\subsection{Visualization and Analysis of EEG–Visual Similarity}
\label{a_sim_all}
\begin{figure}[H]
  \vskip 0.2in
  \begin{center}
    \centerline{\includegraphics[width=0.9\textwidth]{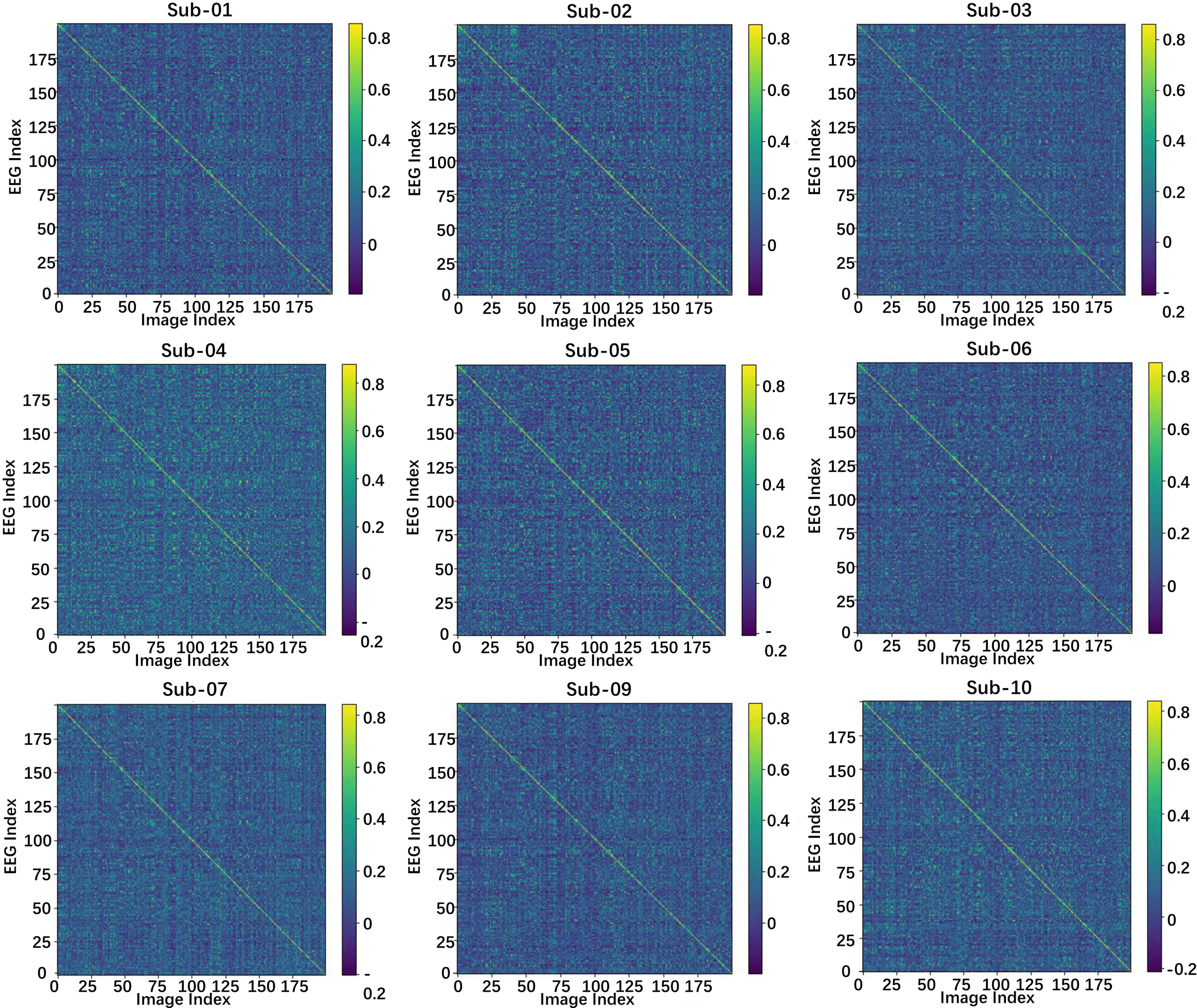}}
    \caption{Similarity matrices for all subjects except Subject 8}
    \label{sim_all}
  \end{center}
\end{figure}
Figure~\ref{sim_all} illustrates the EEG–image similarity matrices for all subjects (excluding Subject 8) under the intra-subject setting. Across all subjects, a prominent diagonal-dominant structure is clearly observed, indicating a strong one-to-one correspondence between the learned EEG representations and their corresponding visual stimuli. In contrast, the off-diagonal values consistently remain at a low level and do not exhibit block-wise or column-wise patterns, suggesting that the learned representations are highly discriminative and free from systematic bias or shortcut correlations. Notably, despite the substantial inter-subject variability present in the raw EEG signals, the overall similarity structure remains consistent across subjects, demonstrating the robustness and stability of the proposed alignment framework.

It is worth noting that Subject 4 exhibits relatively elevated similarity responses in the off-diagonal regions compared to other subjects, indicating increased ambiguity in cross-concept discrimination. This phenomenon may be attributed to individual differences in neural signal quality, cognitive state, or variability in visual perception and attentional engagement during data acquisition. Such reduced separability in the learned representation space is consistent with the comparatively lower decoding accuracy observed for this subject, further supporting the correspondence between the similarity structure and quantitative performance metrics.

\subsection{Visualization of Top-5 Retrieval Results for Representative Samples}
\label{pictures}
\begin{figure}[H]
  \vskip 0.2in
  \begin{center}
    \centerline{\includegraphics[width=0.8\textwidth]{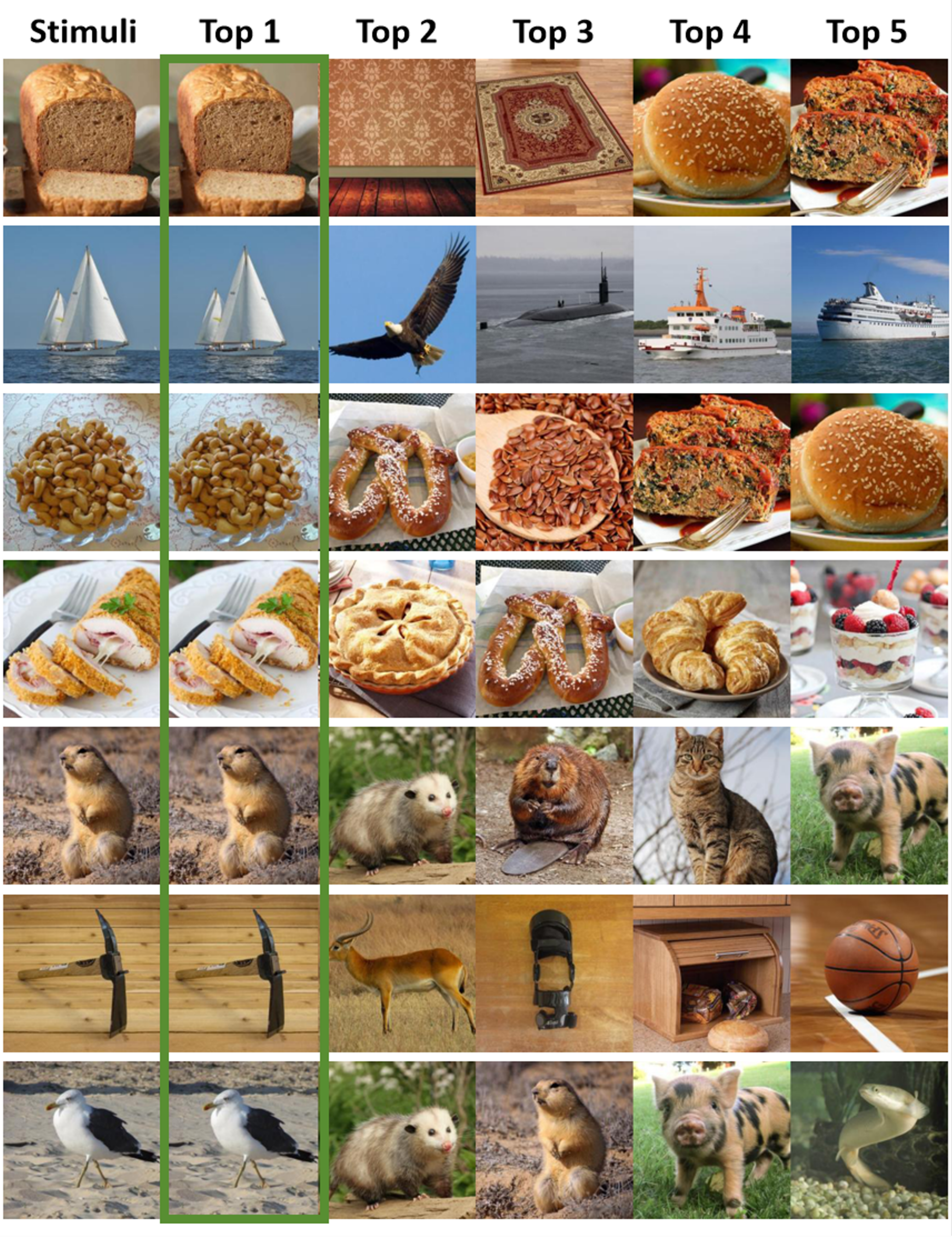}}
    \caption{Good Cases:Top-5 Retrieval Results for Various Stimuli}
    \label{top-5-r}
  \end{center}
\end{figure}

We present qualitative Top-5 retrieval results on the THINGS-EEG dataset, covering both successful and failure cases, as illustrated in Figure~\ref{top-5-r} and Figure~\ref{top-5-w}, respectively. 

In the successful cases, the retrieved images closely match the target stimuli at the semantic level, indicating effective EEG–visual alignment. Notably, the retrieved candidates preserve key visual characteristics of the stimuli, such as overall shape and structural configuration, rather than exact instance-level matching.

In the failure cases, although the Top-1 retrieval is incorrect, semantically related images frequently appear among the Top-5 results, suggesting that the model captures coarse semantic information from EEG signals while finer-grained distinctions remain challenging. These qualitative observations provide further evidence of the robustness of the learned cross-modal representations.
\begin{figure}[H]
  \vskip 0.2in
  \begin{center}
    \centerline{\includegraphics[width=0.8\textwidth]{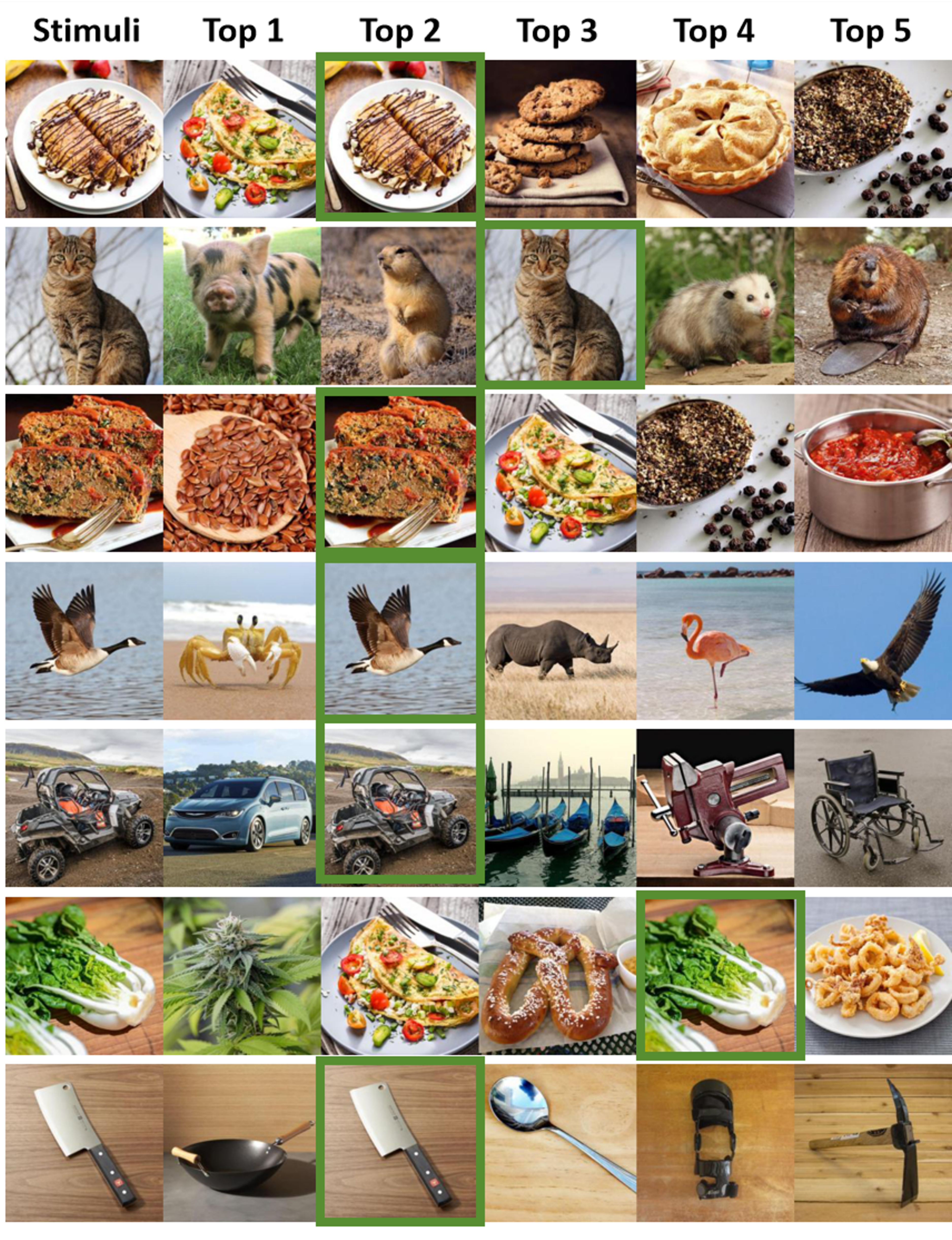}}
    \caption{Bad Cases:Top-5 Retrieval Results for Various Stimuli}
    \label{top-5-w}
  \end{center}
\end{figure}
\newpage

\end{document}